\documentclass[sigconf,nonacm]{acmart}

\copyrightyear{2025}
\acmYear{2025}
\setcopyright{cc}
\setcctype{by}                    

\acmConference[MM '25]{Proceedings of the 33rd ACM International Conference on Multimedia}{October 27--31, 2025}{Dublin, Ireland}
\acmBooktitle{Proceedings of the 33rd ACM International Conference on Multimedia (MM '25), October 27--31, 2025, Dublin, Ireland}
\acmDOI{10.1145/3746027.3754863}
\acmISBN{979-8-4007-2035-2/2025/10}

\AtBeginDocument{}

\usepackage{graphicx}
\usepackage{bm}
\usepackage{multirow}
\usepackage{booktabs}
\usepackage{arydshln}
\usepackage{makecell}
\usepackage{xcolor}
\usepackage{colortbl}

\definecolor{colorsim}{RGB}{224,238,238}
\definecolor{colorist}{RGB}{225,240,213}
\definecolor{colorleaf}{RGB}{255,214,165}
\definecolor{colorbaseline}{RGB}{235,235,235}

\usepackage{bibunits}
\defaultbibliographystyle{ACM-Reference-Format}
\defaultbibliography{refs,refs_app}

\begin{document}

\title{\raisebox{-0.4em}{\includegraphics[height=1.5em]{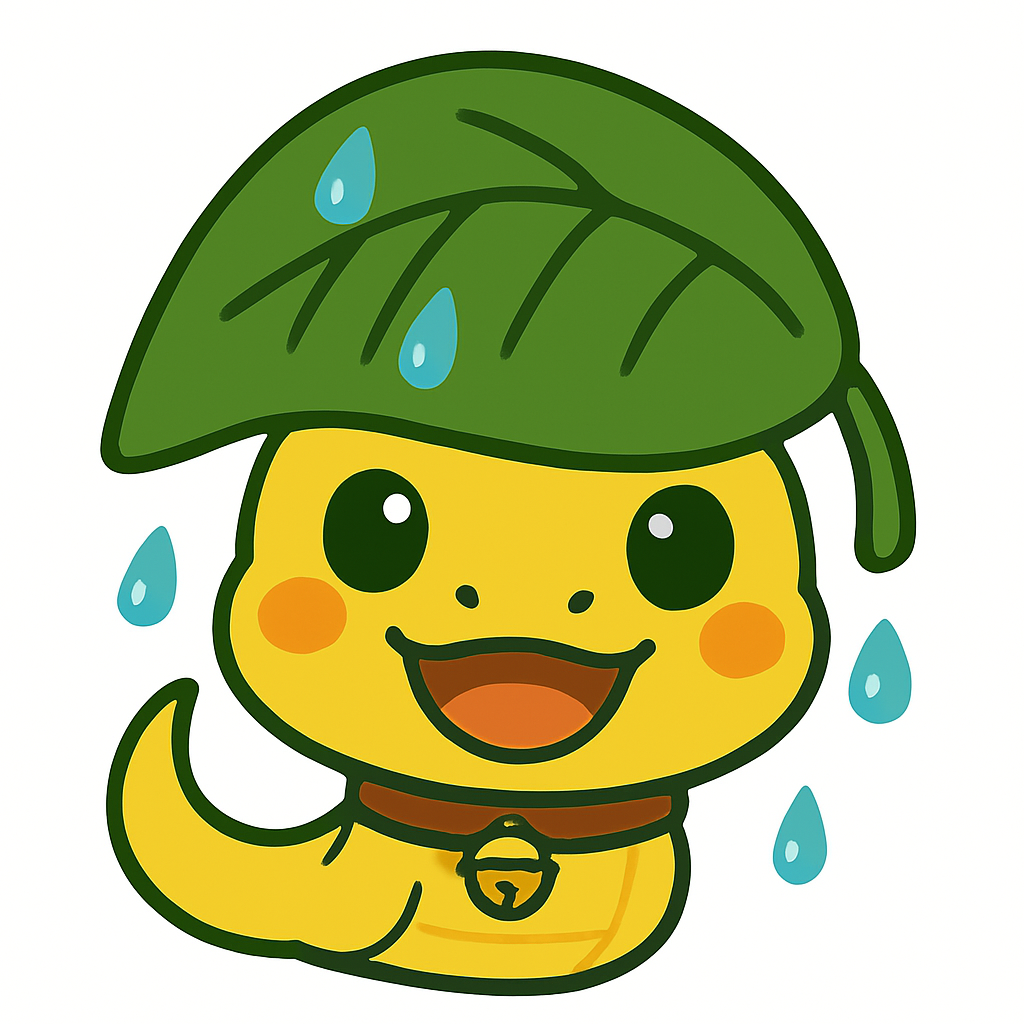}}LEAF-Mamba: Local Emphatic and Adaptive Fusion State Space Model for RGB-D Salient Object Detection}

\author{Lanhu Wu}
\orcid{0009-0006-6420-5971}
\affiliation{%
  \institution{Dalian University of Technology}
  \city{Dalian}
  \country{China}
  }
\email{lanhoong0406@gmail.com}

\author{Zilin Gao}
\orcid{0000-0003-1757-9349}
\affiliation{%
  \institution{Dalian University of Technology}
  \city{Dalian}
  \country{China}
  }
\email{gzl@mail.dlut.edu.cn}

\author{Hao Fei}
\orcid{0000-0003-3026-6347}
\authornote{Corresponding Author.}
\affiliation{%
  \institution{National University of Singapore}
  \city{Singapore}
  \country{Singapore}
}
\email{haofei37@nus.edu.sg}

\author{Mong-Li Lee}
\orcid{0000-0002-9636-388X}
\affiliation{%
  \institution{National University of Singapore}
  \city{Singapore}
  \country{Singapore}
}
\email{dcsleeml@nus.edu.sg}

\author{Wynne Hsu}
\orcid{0000-0002-4142-8893}
\affiliation{%
  \institution{National University of Singapore}
  \city{Singapore}
  \country{Singapore}
}
\email{dcshsuw@nus.edu.sg}

%
\renewcommand{\shortauthors}{Lanhu Wu, Zilin Gao, Hao Fei, Mong-Li Lee, and Wynne Hsu}

\begin{bibunit}

\begin{abstract}
RGB-D salient object detection (SOD) aims to identify the most conspicuous objects in a scene with the incorporation of depth cues. Existing methods mainly rely on CNNs, limited by the local receptive fields, or Vision Transformers that suffer from the cost of quadratic complexity, posing a challenge in balancing performance and computational efficiency. Recently, state space models (SSM), Mamba, have shown great potential for modeling long-range dependency with linear complexity. However, directly applying SSM to RGB-D SOD may lead to deficient local semantics as well as the inadequate cross-modality fusion. To address these issues, we propose a \textbf{L}ocal \textbf{E}mphatic and \textbf{A}daptive \textbf{F}usion state space model (\textbf{LEAF-Mamba}) that contains two novel components: 1) a local emphatic state space module (LE-SSM) to capture multi-scale local dependencies for both modalities. 2) an SSM-based adaptive fusion module (AFM) for complementary cross-modality interaction and reliable cross-modality integration. Extensive experiments demonstrate that the LEAF-Mamba consistently outperforms 16 state-of-the-art RGB-D SOD methods in both efficacy and efficiency. Moreover, our method can achieve excellent performance on the RGB-T SOD task, proving a powerful generalization ability. 
\end{abstract}





\maketitle

\section{Introduction}

\begin{figure}[!t]
  \centering
  \includegraphics[width=0.98\linewidth]{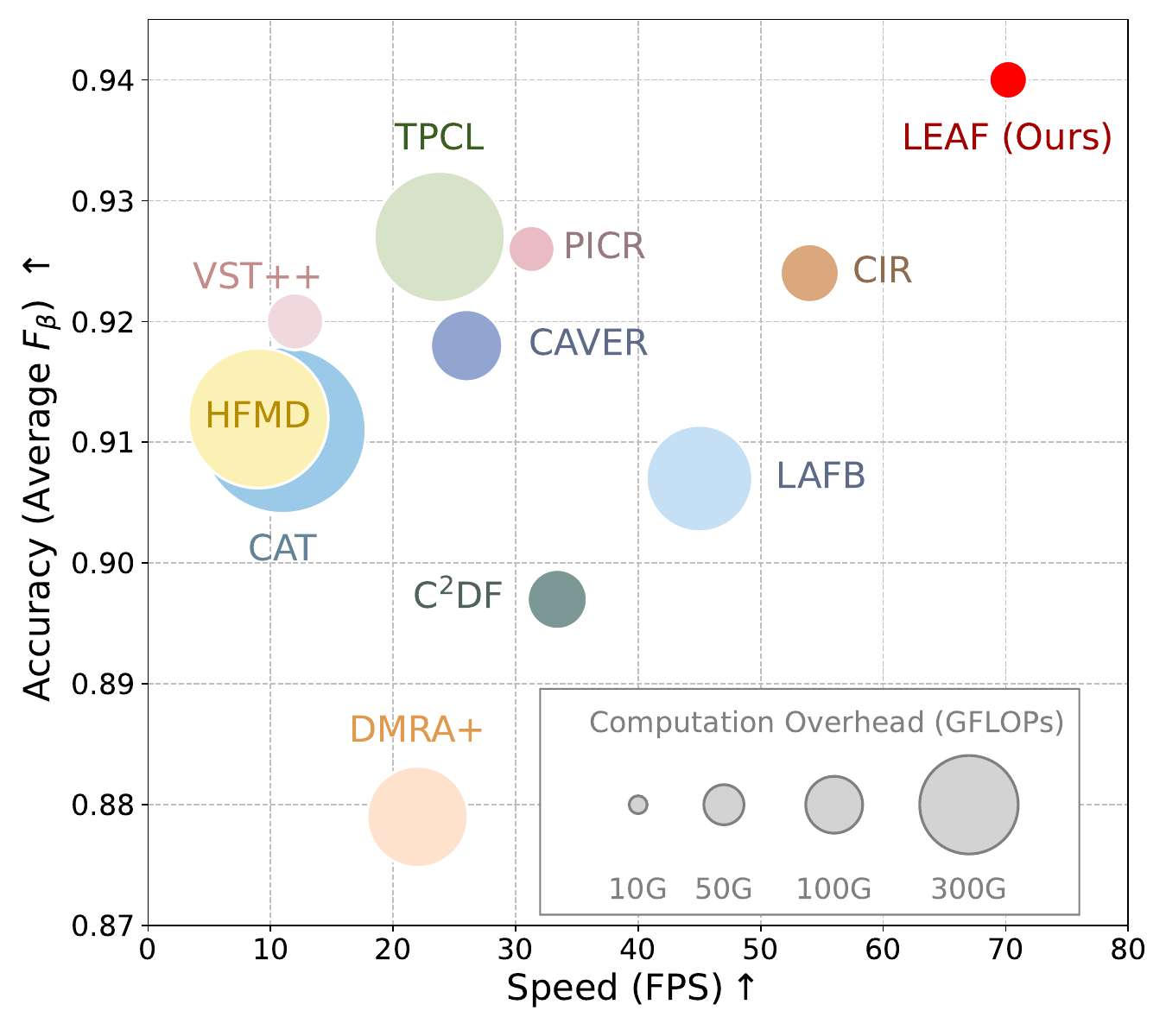}
  \vspace{-2mm}
  \caption{The comparisons with baselines on RGB-D SOD, with respect to efficacy and efficiency. The accuracy indicates the average $F_{\beta}$ on NJUD \cite{ju2014depth}, NLPR \cite{peng2014rgbd} and STERE \cite{niu2012leveraging}.}
  \Description{Task comparisons.}
  \label{fig:intro}
\end{figure}

Salient object detection (SOD) is one of the fundamental vision computing tasks, aiming to pinpoint the most prominent objects in an image. 
Yet RGB SOD \cite{wu2019cascaded, qin2019basnet, wei2020f3net, liu2022poolnet+} may struggle in challenging scenarios such as complex backgrounds and similar appearances between objects and their surroundings.
Thus, depth data, with affluent spatial structure information, is naturally utilized as a supplementary input in addition to the RGB image for accurate saliency prediction, resulting in the task of RGB-D SOD \cite{mahadevan2009saliency,zhao2019contrast}.
Numerous prior methods have been proposed for RGB-D SOD.
Early methods predominantly rely on the convolutional neural networks (CNNs) for single-modality representation and cross-modality fusion, focusing on discriminative modeling \cite{chen2020progressively, zhang2020asymmetric}, feature fusion \cite{fan2020bbs, li2020cross, yao2023depth, hu2024cross}, information optimization \cite{chen2020improved, ji2021calibrated}, model lightweighting \cite{piao2020a2dele, zhang2021depth, wu2022mobilesal}. 
However, CNN-based methods inherently suffer from the limited receptive field of convolutional operation, posing challenges for capturing long-range dependencies. 
To overcome this, a series of Transformer-based methods \cite{cong2023point, sun2023catnet, chen2024disentangled} are developed, leveraging the self-attention mechanism \cite{vaswani2017attention} for global context modeling, thus achieving the state-of-the-art (SoTA).  
Nevertheless, Transformers can be constrained by high computational complexity due to the quadratic growth of resources with the increase in tokens, sacrificing efficiency. 
While some attempts \cite{liu2021tritransnet, pang2023caver} improve the efficiency by reducing the dimension of processing features, they compromise the extent of the receptive fields.  
Therefore, achieving \textbf{\emph{high performance}} meanwhile maintaining \textbf{\emph{model efficiency}} becomes the key bottleneck of RGB-D SOD.
Figure \ref{fig:intro} compares the existing RGB-D SOD research in these two dimensions.

Recently, the newly merged state space models (SSMs), especially Mamba \cite{gu2023mamba}, have shown great potential for modeling long-range dependency with linear complexity, achieving excellent performance with prominent efficiency advantage in various visual tasks \cite{wu2023next,fei2024vitron,wu2024tokenization,fei2024enhancing,fei2025path,wu2025universal}, such as image classification \cite{vim, liu2024vmamba}, video analysis \cite{li2024videomamba} and pathological diagnosis \cite{nasiri2024vim4path, yang2024mambamil, wu2024cnn, wu2024edge}. 
One may directly integrate SSM backbone for RGB-D SOD, by modeling the 2D selective scan (SS2D) mechanism to process vision data, i.e., VMamba \cite{liu2024vmamba}. 
Despite bridging 1D array scanning and 2D plane traversal, it struggles with maintaining the proximity of adjacent tokens, which is critical for local representation modeling in RGB-D SOD. 
While the windowed selective scan strategy might be helpful \cite{huang2024localmamba, wang2024local}, it still fails to capture the multi-scale information due to the fixed window size. 
Besides, current SSM-based methods tend to treat the RGB and depth features equally during the process of intermediate fusion \cite{wan2024sigma, gao2025msfmamba}, while, unfortunately, they largely ignore the cross-modality complementarity and single-modality reliability evidently.

\begin{figure*}[t]
  \centering
  \includegraphics[width=0.99\linewidth]{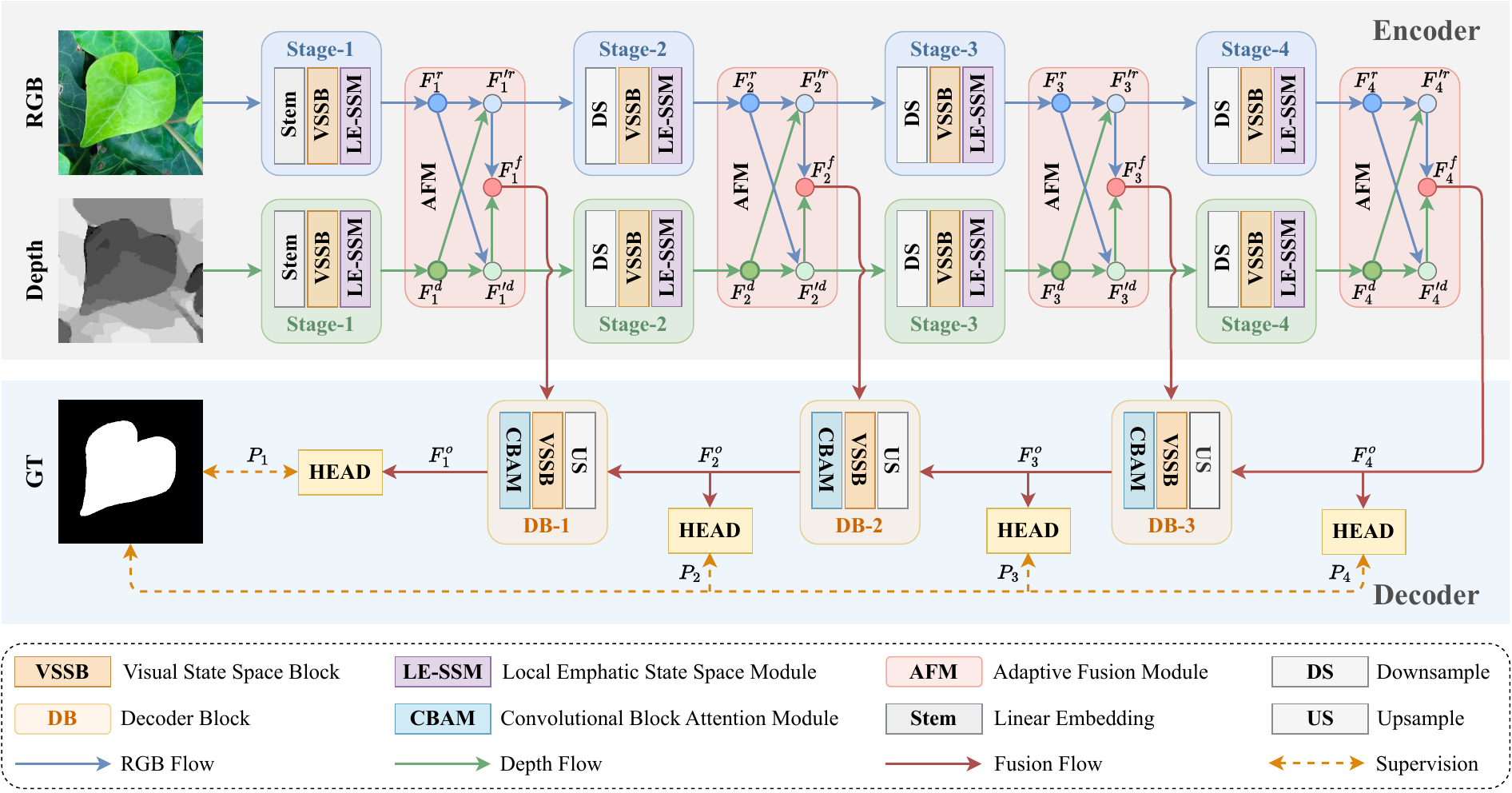}
  \vspace{-1mm}
  \caption{The whole pipeline of our proposed LEAF-Mamba, which consists of two main components: local emphatic state space module (LE-SSM) and adaptive fusion module (AFM). Please refer to Section~\ref{Methodology} for details.}
  \Description{The overview of LEAF.}
  \label{fig:overview}
\end{figure*}

This work is dedicated to addressing both the efficacy and efficiency bottlenecks in existing RGB-D SOD.
We introduce a novel \emph{Local Emphatic and Adaptive Fusion} SSM system, namely \textbf{LEAF-Mamba}, as shown in Figure \ref{fig:overview}.
\textbf{First}, a local emphatic state space module (LE-SSM) is proposed to enrich multi-scale local information in intermediate features of each modality via the multi-scale windowed 2D selective scan (MSW-SS2D). 
Different from existing SS2D \cite{liu2024vmamba}, our MSW-SS2D adopts a four-scale windowed scanning mechanism in four ways for spatial domain traversal. 
As such, adjacent tokens are fully aggregated in multi-scale windows for local modeling without extra computational cost. 
\textbf{Second}, an SSM-based adaptive fusion module (AFM) is devised for cross-modality interaction and integration at multiple stages. 
Specifically, the AFM incorporates a cross-modality second-order pooling (CSoP) layer to compute the modality-specific similarity between RGB and depth features. 
Based on this, the AFM selects the discriminative regions for cross-modality interaction as well as the similar regions for cross-modality fusion under the paradigm of SSM. 
In this way, our system achieves complementary interplay and reliable integration of two modalities in an attentive manner. 
Attributing to the comprehensive feature representation, robust cross-modality fusion and efficient SSM, our LEAF-Mamba delivers great performance with a relatively low computational cost, as depicted in Figure~\ref{fig:intro}. 

Experimentally, we validate our method on seven RGB-D SOD benchmarks including NJUD \cite{ju2014depth}, NLPR \cite{peng2014rgbd}, SIP \cite{fan2020rethinking}, STERE \cite{niu2012leveraging}, SSD \cite{zhu2017three}, LFSD \cite{li2014saliency} and DUT-D \cite{piao2019depth}, where the results demonstrate its superiority over 16 SoTA methods in terms of both efficacy and efficiency. 
Particularly, LEAF-Mamba reduces the MAE by 13.2\% and 8.16\% on SSD and LFSD with only 18.1 GFLOPs and an astonishing real-time speed of 70.2 FPS. 
Besides, we conduct detailed ablations to verify the effectiveness of the proposed LE-SSM and AFM in multi-scale local enhancement and selective cross-modality fusion, respectively. 
Moreover, we extend our method to the RGB-T SOD task and further demonstrate its prominent generalizability. 

To sum up, in this paper we propose a novel RGB-D SOD system (\textbf{LEAF-Mamba}) based on the SSM technique, where our main contributions are threefold:
\begin{itemize}
\item We devise a novel local emphatic state space module (LE-SSM) which performs a four-scale windowed selective scan to enrich multi-scale local information with low computational cost.
\item We introduce an SSM-based adaptive fusion module (AFM) with a modality-specific selective mechanism, which dynamically interacts the complementary cues and fuses the reliable content in RGB and depth features.
\item Our system not only sets new records on 7 RGB-D SOD benchmarks, but also achieves a low computation overhead of 18.1 GFLOPs and a real-time inference speed of 70.2 FPS. Also, it shows prominent generalizability on RGB-T SOD task.
\end{itemize}

\section{Related Work}

\subsection{RGB-D Salient Object Detection}
RGB-D SOD 
combines RGB images with depth cues to identify the most conspicuous objects in a scene.
With the development of deep learning technology, CNN-based methods are firstly proposed for RGB-D SOD.
Zhang~\emph{et~al.}~\cite{zhang2020select} introduce a complimentary interaction module to discriminatively select useful representation from the RGB and depth data.
Ji~\emph{et~al.}~\cite{ji2021calibrated} propose a depth calibration and fusion framework to calibrate the depth map and realize an efficient fusion of two modalities. 
Wu~\emph{et~al.}~\cite{wu2022mobilesal} present an efficient RGB-D SOD method based on mobile network and an implicit depth restoration technique to strengthen the mobile backbones.

However, due to the confined receptive field of CNN, these methods are deficient in extracting global context.
To this end, some Transformer-based methods are proposed, which realize long-range modeling by self-attention mechanism. 
Liu~\emph{et~al.}~\cite{liu2021tritransnet} introduce a triplet Transformer embedding module for modeling
high-level features. 
Pang~\emph{et~al.}~\cite{pang2023caver} propound a view-mixed Transformer to excavate the global cues in intra-modal features and simplify the cross-modal interaction and alignment. 
Cong~\emph{et~al.}~\cite{cong2023point} set forth a CNN-assisted Transformer architecture with point-aware interaction and CNN-induced refinement. 
Despite their promising results, these methods typically suffer from the quadratic scaling inherent in the self-attention mechanism, particularly for the dual-modality scenario \cite{chen-etal-2024-m3cot,fei2024video,chen2025towards,wang2025multimodal,cheng2025comt}. Different from them, our proposed method benefits from the linear overhead of Mamba, making a trade-off between effectiveness and efficiency, as shown in Figure~\ref{fig:intro}.

\subsection{State Space Models}
State space models (SSMs) \cite{gu2021efficiently, smith2022simplified}, with linear complexity, have emerged as compelling alternatives to Transformers for modeling long-range dependency. 
Recently, Gu~\emph{et~al.}~\cite{gu2023mamba} propose the selective state space model, Mamba, which demonstrates superior performance over Transformers in NLP. Inspired by its remarkable performance, researchers extend it to the domain of computer vision. Zhu~\emph{et~al.}~\cite{vim} integrate SSM with bidirectional scanning, making each patch related to another. Liu~\emph{et~al.}~\cite{liu2024vmamba} extend the scanning in both horizontal and vertical directions to further interpret spatial relationships. However, they elongate the distance between adjacent tokens, overlooking the preservation of local 2D dependency. 
To this end, Huang~\emph{et~al.}~\cite{huang2024localmamba} introduce a local scanning strategy that divides images into distinct windows to capture local dependencies while maintaining a global perspective.
Despite its effectiveness, it relies on the fixed-size window for local modeling, limiting the diversity of local dependencies. 
In contrast, our proposed LE-SSM adopts a four-way four-scale windowed scanning strategy, which enriches the multi-scale local information.

SSMs have been preliminarily employed and explored in a wide range of multi-modal tasks. Wan~\emph{et~al.}~\cite{wan2024sigma} introduce a Siamese Mamba network for multi-modal semantic segmentation with a fusion module.
Gao~\emph{et~al.}~\cite{gao2025msfmamba} propose a multi-scale feature fusion Mamba for multi-source remote sensing image classification.
However, they treat dual modalities equally during the cross-modality fusion, ignoring their complementarity and reliability.
Conversely, our SSM-based AFM selectively interacts with the complementary cues and fuses the reliable messages in RGB and depth features.

\section{Methodology}
\label{Methodology}
In this section, we first introduce some essential concepts of the state space model.
Then, we provide a detailed description of our LEAF-Mamba, including its overall framework and module design.
Figure \ref{fig:overview} illustrates the overall architecture of LEAF-Mamba.

\subsection{Preliminaries}
\noindent \textbf{State Space Models.}
SSM is a linear time-invariant system that maps an input sequence 
\(\displaystyle x(t) \in \mathbb{R}^N\) to an output sequence 
\(\displaystyle y(t) \in \mathbb{R}^N\). They are mathematically 
represented by the following linear ordinary differential equations:
\begin{equation}
    \begin{aligned}
  h'(t) &= \bm{A} h(t) + \bm{B} x(t), \\
  y(t) &= \bm{C} h(t) + \bm{D}  x(t), 
\end{aligned}
\end{equation}
where \(h(t) \in \mathbb{R}^N\) indicates a hidden state, 
\(h'(t) \in \mathbb{R}^N\) refers to the time derivative of \(h(t)\), 
and \(N\) is the number of states. Additionally, 
\(\bm{A} \in \mathbb{R}^{N \times N}\) is the state transition matrix, 
\(\bm{B} \in \mathbb{R}^{N \times 1}\), \(\bm{C} \in \mathbb{R}^{1 \times N}\) are projection matrices, and \(\bm{D} \in \mathbb{R}^{N \times 1}\) is a residual connection.

SSMs are continuous-time models, and are challenging to incorporate into deep learning networks. To address this, discrete 
versions of SSMs are proposed. The ordinary differential equations are 
discretized by the zero-order hold rule. A timescale parameter \(\Delta\) 
is introduced to convert the continuous parameters \(\bm{A}\) and \(\bm{B}\) into 
discrete parameters \(\overline{\bm{A}}\) and \(\overline{\bm{B}}\), respectively, 
as:
\begin{equation}
    \begin{aligned}
  \overline{\bm{A}} &= \exp(\Delta \bm{A}), \\
  \overline{\bm{B}} &= (\Delta \bm{A})^{-1} (\exp(\Delta \bm{A}) - \bm{I}) \cdot \Delta \bm{B} \approx \Delta \bm{B}. 
\end{aligned}
\end{equation}
The matrix \(\overline{\bm{B}}\) can be approximated by applying a first-order Taylor expansion to the term involving the matrix exponential. After discretization, the SSM system can be reformulated as:
\begin{equation}
    \begin{aligned}
  h_t &= \overline{\bm{A}} h_{t-1} + \overline{\bm{B}} x_t, \\
  y_t &= \bm{C} h_t + \bm{D} x_t.
\end{aligned}
\end{equation}
Further, the models compute output through a global convolution.
\begin{equation}
\begin{aligned}
\overline{\bm{K}} &= (\bm{C}\overline{\bm{B}}, \bm{C}\overline{\bm{A}} \overline{\bm{B}}, \ldots, \bm{C}\overline{\bm{A}}^{L-1}\overline{\bm{B}}), \\
y &= x * \overline{\bm{K}},
\end{aligned}
\end{equation}
where $L$ is the length of the input sequence $x$, and $\overline{\bm{K}} \in \mathbb{R}^L$ is a structured convolutional kernel.

\vspace{1mm}
\noindent \textbf{Selective Scan Mechanism.} Traditional SSMs adopt a linear time-invariant framework, wherein the projection matrices remain fixed and unaffected by variations in the input sequence. However, this static configuration results in a lack of attention on individual elements within the sequence. To overcome this limitation, Mamba \cite{gu2023mamba} introduces a selective scan mechanism where the parameter matrices become input-dependent. In this way, SSMs can better model the complex interactions present in long sequences through the transformation into linear time-varying systems.

\subsection{Overview of LEAF-Mamba}
The overall framework of our proposed LEAF-Mamba is shown in Figure~\ref{fig:overview}, which follows a standard encoder-decoder architecture. The encoder is a dual-stream structure built upon VMamba~\cite{liu2024vmamba}, which yields multi-stage features from RGB and depth images. To enrich multi-scale local semantics, the last VMamba block of each stage is substituted with our local emphatic state space module (LE-SSM), resulting in local-enhanced dual-modality features $\{F^r_i\}^4_{i=1}$ and $\{F^d_i\}^4_{i=1}$. Then, they are fed into the adaptive fusion module (AFM) for attentive cross-modality interaction and integration. Specifically, AFM communicates the complementary cues in $\{F^r_i\}^4_{i=1}$ and $\{F^d_i\}^4_{i=1}$ for inter-enhanced features $\{F'^r_i\}^4_{i=1}$ and $\{F'^d_i\}^4_{i=1}$ which act as the inputs for next stage. Meanwhile, AFM fuses the reliable content of $\{F'^r_i\}^4_{i=1}$ and $\{F'^d_i\}^4_{i=1}$ for RGB-D features $\{F^f_i\}^4_{i=1}$. The multi-stage RGB-D features are put into an SSM-based FPN \cite{lin2017feature} decoder for multi-level predictions $\{P_i\}^4_{i=1}$ with deep supervision, where the CBAM \cite{woo2018cbam} is adopted to facilitate the spatial- and channel-wise variation of the processing features. We take $P_1$ as the final prediction map.

\subsection{Local Emphatic State Space Module}
Early vision mamba methods \cite{vim, liu2024vmamba, yang2024plainmamba} always flatten 2D plane into 1D array along rows and columns, disrupting the proximity of adjacent tokens. Although recent works \cite{huang2024localmamba, wang2024local, dang2024log} adopt the windowed scan strategy for local modeling, they fail to capture the multi-scale semantic cues due to the fixed window size. To solve this problem, we propose a local emphatic state space module (LE-SSM), which captures multi-view local dependencies via a multi-scale windowed 2D selective scan (MSW-SS2D) mechanism.

\begin{figure}[t]
  \centering
  \includegraphics[width=\linewidth]{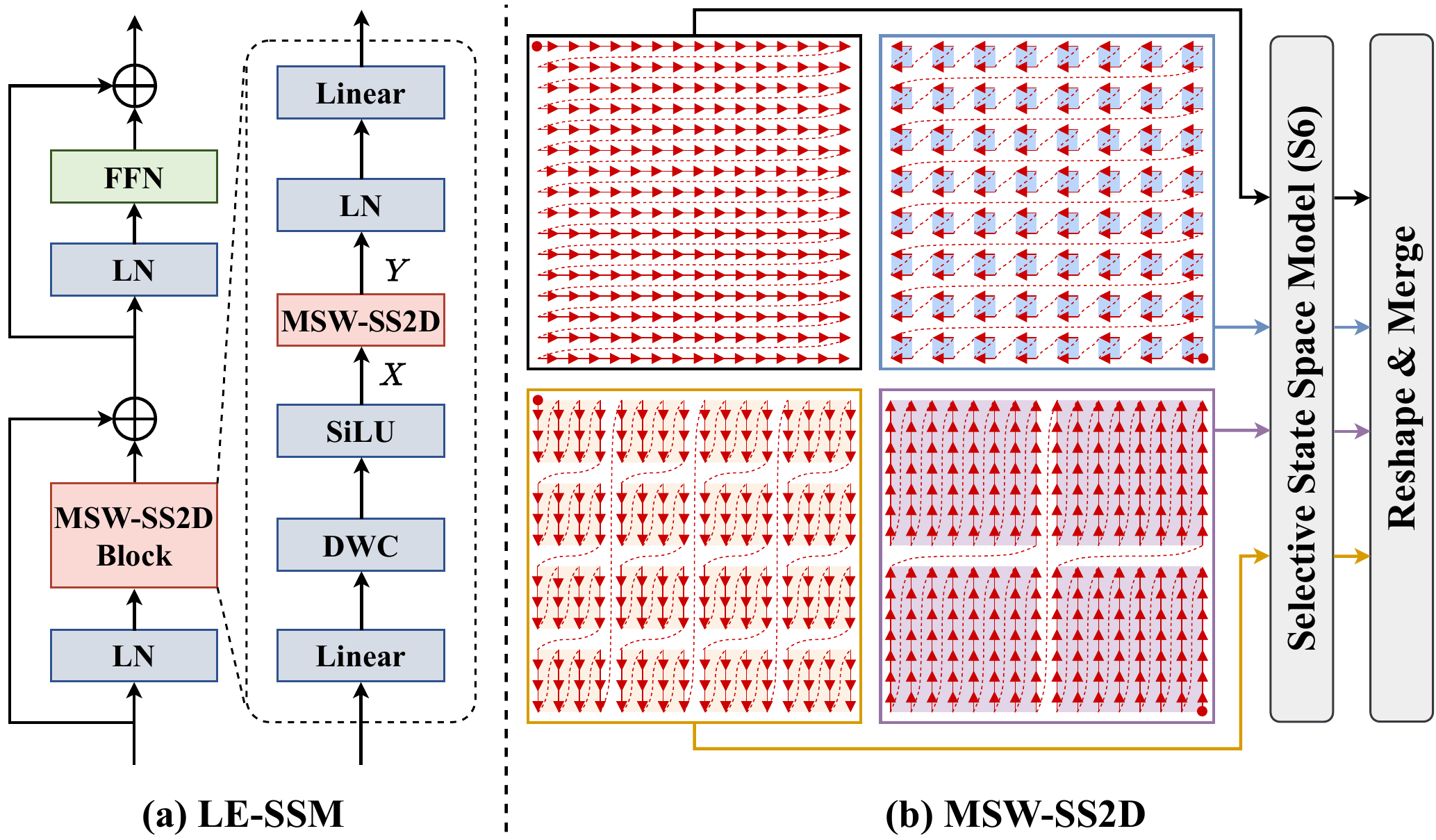}
  \caption{Illustration of the local emphatic state space module (LE-SSM) and multi-scale windowed 2D selective scan (MSW-SS2D) mechanism.}
  \Description{The structure of MSW-SS2D.}
  \label{fig:LE-SSM}
  \vspace{-1mm}
\end{figure}

The illustration of LE-SSM is provided in Figure~\ref{fig:LE-SSM}~(a). Unlike the Mamba \cite{gu2023mamba} used in NLP, the LE-SSM consists of a single network branch with two residual modules, mimicking the architecture of Transformer block \cite{dosovitskiy2020image}. Meanwhile, the S6 block of Mamba is substituted with the newly proposed multi-scale windowed 2D selective scan (MSW-SS2D), shown in Figure~\ref{fig:LE-SSM} (b). Specifically, given an input as $X$, MSW-SS2D first unfolds it into sequences along four distinct traversal paths, i.e., horizontal (H) and vertical (V) directions along with their flipped counterparts (HF and VF). Each of them adopts the windowed selective scan strategy with a unique window size. In this paper, the standard/flipped horizontal scan adopts the window size of 1/2 while the standard/flipped vertical scan adopts the window size of 4/8, denoted as $\mathrm{H_1}$, $\mathrm{HF_2}$, $\mathrm{V_4}$ and $\mathrm{VF_8}$. Notably, the matching between directions and window sizes can be set randomly because the rotation augmentation during the training phase can make them fully connected. Then each sequence is processed in parallel using a separate S6 block, and the resultant sequences are reshaped and merged to form the output $Y$. The whole procedure of MSW-SS2D can be formulated as:
\begin{equation}
    Y = {\sum}_{\mathrm{Scan \in \mathcal{S}}} \, \text{Reshape}(\text{S6}(\text{Scan}(X))),
\end{equation}
where $\mathcal{S}$ = $\mathrm{\{H_1, HF_2, V_4, VF_8\}}$, denoting the set of traversal paths. 

In this case, MSW-SS2D achieves a four-way, four-scale windowed selective scan, which enables the LE-SSM to effectively extract the multi-scale local information without extra computational cost compared to SS2D \cite{liu2024vmamba}. 

\begin{figure*}[t]
  \centering
  \includegraphics[width=0.98\linewidth]{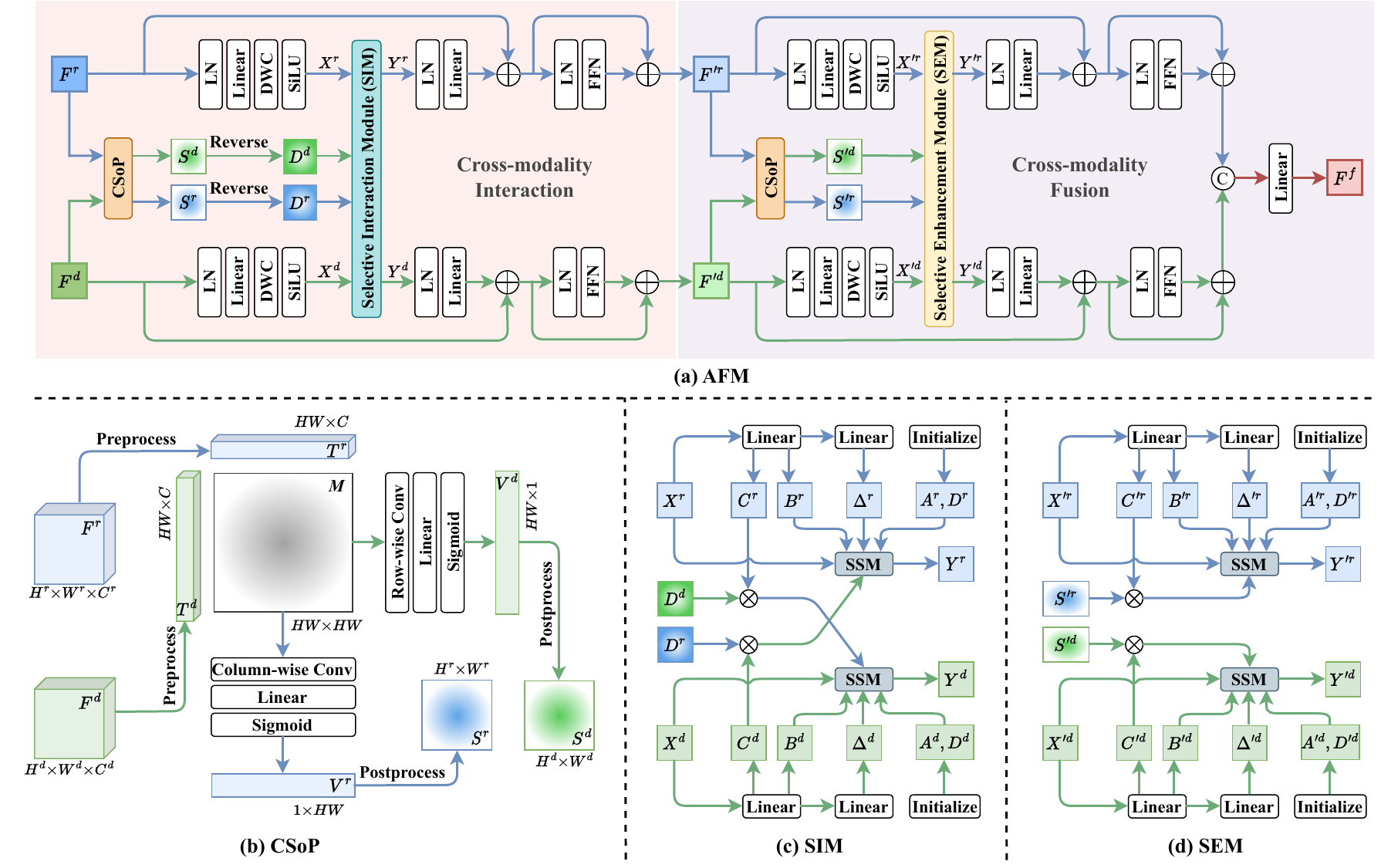}
  \vspace{-1mm}
  \caption{Illustration of the adaptive fusion module (AFM) and its components.}
  \Description{The structure of AFM.}
  \label{fig:AFM}
\end{figure*}

\subsection{Adaptive Fusion Module}
Current SSM-based multi-modal methods \cite{wan2024sigma, gao2025msfmamba} typically treat the cross-modality features equally during the fusion process, which overlooks their complementarity and reliability. To this end, we design an adaptive fusion module (AFM) to dynamically interact with the complementary cues and fuse the reliable content in RGB and depth features under the paradigm of SSM.

The whole structure of AFM is illustrated in Figure~\ref{fig:AFM}~(a), consisting of two sequential processes: cross-modality interaction and cross-modality fusion. Both of them follow the architecture of VMamba block. For cross-modality interaction, the same-stage RGB and depth features $F^r$ and $F^d$ are first fed into a cross-modality second-order pooling (CSoP) layer to compute the modality-specific similarity maps $S^r$ and $S^d$. Then they are reversed to produce the distance maps $D^r$ and $D^d$, which are utilized to select the discriminative regions of the processing RGB and depth features $X^r$ and $X^d$ for complementary cross-modality interaction in the selective interaction module (SIM). For cross-modality fusion, the interacted dual-modality features $F'^r$ and $F'^d$ are also fed into a CSoP layer to measure their similarity $S'^r$ and $S'^d$. Then, a selective enhancement module (SEM) is incorporated to weight the similar regions for subsequent reliable cross-modality fusion $F^f$. We now depict the technical details of CSoP, SIM and SEM, respectively.

\vspace{1mm}
\noindent \textbf{Cross-modality Second-order Pooling.} The structure of CSoP is illustrated in Figure~\ref{fig:AFM}~(b). Take the CSoP in cross-modality interaction as an example. We first preprocess the cross-modality features $F^r$ and $F^d$ with $1\times1$ convolution ($\mathrm{C}_{1\times1})$ and downsampling (DS) for reducing the number of their channels and spatial size to a fixed $H \times W \times C$, so as to decrease the computational cost of the following operations. Then we flatten them into tokens $T^r$ and $T^d$ along the spatial dimension and compute their covariance matrix $M$. The above process can be formulated as:
\begin{equation}
\begin{aligned}
      T^r &= {\mathrm{Flatten}}({\mathrm{DS}}({\mathrm{{C_{1\times1}}}(F^r)})), \\
     T^d &= {\mathrm{Flatten}}({\mathrm{DS}}({\mathrm{{C_{1\times1}}}(F^d)})), \\
     M_{i,j} &= \mathrm{Cov}(T^d_i, T^r_j).
\end{aligned}
\end{equation}
After that, we separately perform row-wise convolution ($\mathrm{C^{row}}$) and column-wise convolution ($\mathrm{C^{col}}$) on the covariance matrix $M$ for modality-specific pooling. Each of them is followed by a linear layer and a sigmoid function ($\sigma$) for channel scaling and nonlinear activation respectively, outputting the weighted vectors $V^r$ and $V^d$ of length $HW$, formulated as:
\begin{equation}
\begin{aligned}
      V^r &= \sigma(\mathrm{Linear}(\mathrm{C^{col}}(M))), \\
      V^d &= \sigma(\mathrm{Linear}(\mathrm{C^{row}}(M))).
\end{aligned}
\end{equation}
Finally, we postprocess the weighted vectors $V^r$ and $V^d$ by reshaping them into $H\times W$ matrices and upsampling (US) them to the original spatial size for modality-specific similarity $S^r$ and $S^d$, written as:
\begin{equation}
\begin{aligned}
      S^r &= \mathrm{US}(\mathrm{Reshape}(V^r)), \\
      S^d &= \mathrm{US}(\mathrm{Reshape}(V^d)).
\end{aligned}
\end{equation}

As such, our CSoP extends the traditional point-to-point similarity to a points-to-points manner, which further explores the relevance of cross-modality features from a global perspective.

\newcommand{\tablestyle}[2]{\setlength{\tabcolsep}{#1}\renewcommand{\arraystretch}{#2}\centering\footnotesize}
\newcommand{\ourfont}[1]{\footnotesize{#1}}

\newcommand{\lineheight}{1.2}

\begin{table*}[t]
\tablestyle{3.9pt}{1.05}
\caption{Quantitative comparisons with state-of-the-arts on seven benchmark datasets. $\uparrow$ ($\downarrow$) denotes higher the better (lower the better). The best and second-best results are shown in \textcolor{red}{Red} and \textcolor{blue}{Blue}, respectively. $-$ means not available.}
\vspace{-1mm}
\centering
\label{table:compare with sota}
\begin{tabular}{lr|ccccccc|ccccccccc|c}
\hline
\multicolumn{2}{c}{Backbone} & \multicolumn{7}{|c}{CNN} & \multicolumn{9}{|c|}{Transformer} & SSM\\

\hline
\multicolumn{2}{c|}{\multirow{2}{*}{Method}}  &  \ourfont{DMRA+} & \ourfont{C2DF} & \ourfont{DCMF} & \ourfont{CIR} & \ourfont{DIF} & \ourfont{MFUR} & \ourfont{LAFB}  & \ourfont{MITF} & \ourfont{PICR}& \ourfont{CAVER} & \ourfont{CAT} & \ourfont{TPCL} & \ourfont{HFMD} & \ourfont{EM-T} & \ourfont{VST++} &\ourfont{DCT} & \ourfont{LEAF} \\

& & \cite{ji2022dmra} & ~\cite{C2DF} & \cite{DCMF} & \cite{cong2022cir} & \cite{yao2023depth} & ~\cite{MFUR} & \cite{wang2024local} & \cite{MITF} & \cite{cong2023point} & \cite{pang2023caver} & \cite{sun2023catnet} & \cite{TPCL} & \cite{HFMD} & \cite{chen2024trans} & \cite{liu2024vst++} & \cite{DCT} &$-$ \\

\multicolumn{2}{c|}{Publish} & \ourfont{TIP} & \ourfont{TMM} & \ourfont{TIP} & \ourfont{TIP} & \ourfont{TIP} & \ourfont{KBS} & \ourfont{TCSVT} & \ourfont{TCSVT} & \ourfont{MM} & \ourfont{TIP} & \ourfont{TMM} & \ourfont{TMM} & \ourfont{TIM} & \ourfont{TNNLS} & \ourfont{TPAMI} & \ourfont{TIP} & \ourfont{(Ours)} \\
\multicolumn{2}{c|}{Year}   & 2022 & 2022 & 2022 & 2022 & 2023 & 2024 & 2024 & 2022 & 2023 & 2023 & 2023 & 2023 & 2024 & 2024 & 2024 & 2024 & -- \\

\hline
\multirow{4}{*}{\rotatebox{90}{\textit{NJUD}}} &$F_{\beta}\uparrow$  & .882     &.899    &.915    &.928      &.906     &\textcolor{blue}{\textbf{.937}} &.919         &.926          &.931    &.923          &.929	&.930 &.923	&.935       &.927  &.934 &\textcolor{red}{\textbf{.945}}\\
 & $S_{\alpha }\uparrow$ & .905    &$-$     &.913    &.925        &$-$    &.920   &$-$     	    &.923         &.927    &.920                   &\textcolor{blue}{\textbf{.937}}    &.926 &.927    &.931           &.926  &.932 &\textcolor{red}{\textbf{.940}}\\
& $E_{\xi }\uparrow$ & .914   &.919     &.948    &$-$         &.923   &$-$  &.924              &.957        &$-$    &.951                     &.933   &.959 &.956   &\textcolor{blue}{\textbf{.961}}     &.957  &.959 &\textcolor{red}{\textbf{.967}}\\
&$M\downarrow$& .044   &.038     &.043    &.035         &.037  &.035 &.028 	    &.030        &.029    &.031                    &\textcolor{red}{\textbf{.025}}    &.028 &.028  &\textcolor{blue}{\textbf{.027}}	    &.031  &.031 &\textcolor{red}{\textbf{.025}}\\
\hline

\multirow{4}{*}{\rotatebox{90}{\textit{NLPR}}} & $F_{\beta}\uparrow$ & .880    &.899    &.906    &.924          &.906    &.930  &.905             &.928         &.928     &.921                 &.916	&.930  &.913	&\textcolor{blue}{\textbf{.934}}      &.922  &.923 &\textcolor{red}{\textbf{.939}}\\
& $S_{\alpha }\uparrow$ & .926    &$-$     &.922    &.933             &$-$    &.931 &$-$  	    &.933     &.935     &.929                &.939		&.936  &.931     &\textcolor{blue}{\textbf{.940}}        &.934  &.934 &\textcolor{red}{\textbf{.945}}\\
& $E_{\xi }\uparrow$ & .952   &.958     &.954    &$-$             &.960   &$-$  &.958             &.968     &$-$     &.961                   &.968		&\textcolor{blue}{\textbf{.970}}  &.964   &\textcolor{blue}{\textbf{.970}}       &.966 &.965 &\textcolor{red}{\textbf{.976}}\\
&$M\downarrow$& .026    &.021     &.029    &.023        &.020  &.022  &.021    	    &.018     &.019  &.022                   &.018  &\textcolor{blue}{\textbf{.017}}  &.021     &\textcolor{blue}{\textbf{.017}}	   &.020  &.023 &\textcolor{red}{\textbf{.016}}\\

\hline
\multirow{4}{*}{\rotatebox{90}{\textit{SIP}}} & $F_{\beta}\uparrow$ & .863    &$-$    &$-$    &.896       &.873      &.910  &.902               &.913     &$-$       &.906                   &.918	&\textcolor{blue}{\textbf{.922}} &.896	&.920     &.917 &.910 &\textcolor{red}{\textbf{.935}}\\  
&  $S_{\alpha }\uparrow$ &  .852    &$-$     &$-$    &.888      &$-$       &.890 &$-$ 	    &.899     &$-$       &.893                     &\textcolor{blue}{\textbf{.913}}	&.902	&.886 &.903         &.903 &.899 &\textcolor{red}{\textbf{.920}}\\
& $E_{\xi }\uparrow$ & .906    &$-$     &$-$    &$-$      &.915   &$-$  &.937    &.940     &$-$  &.933        &.944	&\textcolor{blue}{\textbf{.946}} &.930	&.944    &\textcolor{blue}{\textbf{.946}} &.942 &\textcolor{red}{\textbf{.950}}\\
&$M\downarrow$ & .060    &$-$     &$-$    &.052      &.051      &.049 &.041        &.040     &$-$  &.042                  &\textcolor{blue}{\textbf{.034}}    &.035  &.044   &.039	&.038  &.038 &\textcolor{red}{\textbf{.032}}\\

\hline
\multirow{4}{*}{\rotatebox{90}{\textit{STERE}}}  & $F_{\beta}\uparrow$ & .875    &.892    &.906    &.914       &.894   &.919  &.896            &.910     &.920  &.911          &.902	&.922  &.901	&\textcolor{blue}{\textbf{.926}}       &.911 &.919 &\textcolor{red}{\textbf{.935}}\\
& $S_{\alpha }\uparrow$ &  .903    &$-$     &.910    &917      &$-$ &.920  &$-$  	    &.909     &.921  &.914       &\textcolor{blue}{\textbf{.925}}	&.920 &.900	&\textcolor{blue}{\textbf{.925}}        &.913 &.922 &\textcolor{red}{\textbf{.933}}\\
& $E_{\xi }\uparrow$  & .920    &.927     &.946    &$-$      &.930  &$-$ &.930            &.953     &$-$  &.949        &.935	&\textcolor{red}{\textbf{.960}} &.943	&\textcolor{blue}{\textbf{.958}}        &.952 &.955 &\textcolor{blue}{\textbf{.958}}\\
& $M\downarrow$ & .043    &.038     &.043    &.038      &.036   &.040   &.037         &.034     &.031   &.033         &.030   &.029   &.040   &\textcolor{blue}{\textbf{.028}}	 &.035 &.035 &\textcolor{red}{\textbf{.026}}\\
\hline

\multirow{4}{*}{\rotatebox{90}{\textit{SSD}}}  & $F_{\beta}\uparrow$ & .824 & .848 & .867 & $-$ & $-$ & $-$ & .860 & .862 & $-$ & .854 & $-$ & $-$ & .871 & .875  & \textcolor{blue}{\textbf{.883}} & $-$ & \textcolor{red}{\textbf{.904}}\\
&$S_{\alpha }\uparrow$  & .868 & $-$  & .882 & $-$ & $-$ & -- & $-$  & .877 & $-$ & .874 & $-$ & $-$ &.887 & .885 & \textcolor{blue}{\textbf{.896}} & $-$ & \textcolor{red}{\textbf{.918}}\\
&$E_{\xi }\uparrow$ & .911 & .911 & .921 & $-$ & $-$ & $-$ & .922 & .914 & $-$ & .924 & $-$ & $-$ & .934 & .935 & \textcolor{blue}{\textbf{.944}} & $-$ & \textcolor{red}{\textbf{.953}}\\
& $M\downarrow$ & .049 & .047 & .053 & $-$ & $-$ & $-$ & .041 & .047 & $-$ & .043 & $-$ & $-$ &\textcolor{blue}{\textbf{.038}} & .039 & \textcolor{blue}{\textbf{.038}} & $-$ & \textcolor{red}{\textbf{.033}}\\
\hline

\multirow{4}{*}{\rotatebox{90}{\textit{LFSD}}} & $F_{\beta}\uparrow$ & .861 & .863 & .875 & .883 & .875 & \textcolor{blue}{\textbf{.891}} & $-$ & .876 & .894 & .886 & .884 & .888 & .883 & $-$ & .887 & $-$ & \textcolor{red}{\textbf{.908}}\\
& $S_{\alpha }\uparrow$ & .871 & $-$  & .878 & .875 & $-$  & .863 & $-$ & .874 & .888 & .882 & \textcolor{blue}{\textbf{.894}} & .892 & .880 & $-$ & .888 & $-$ & \textcolor{red}{\textbf{.907}}\\
&$E_{\xi }\uparrow$ & .902 & .883 & .909 & $-$  & .907 & $-$ & $-$ & .911 & $-$ & .921 & .908 & \textcolor{blue}{\textbf{.926}} & .915 & $-$ & .915 & $-$ & \textcolor{red}{\textbf{.934}}\\
& $M\downarrow$ & .069 & .065 & .068 & .068 & .055 & .075 & $-$ & .063 & .053 & .056 & .051 & \textcolor{blue}{\textbf{.049}} & .059 & $-$ & .060 & $-$ & \textcolor{red}{\textbf{.045}}\\

\hline
\multirow{4}{*}{\rotatebox{90}{\textit{DUT-D}}} &$F_{\beta}\uparrow$ & .911 & .934 & .932 & .938 & .940 & .948 & .930 & .934 & .951 & .942 & .951 & \textcolor{blue}{\textbf{.956}} & .951 & $-$ & .948 & .952 & \textcolor{red}{\textbf{.958}}\\
&$S_{\alpha }\uparrow$ & .919 & $-$  & .928 & .932 & $-$  & .943 & $-$  & .937 & .943 & .931 & \textcolor{red}{\textbf{.953}} & .946 & .950 & $-$ & .943 & .948 & \textcolor{blue}{\textbf{.952}}\\
& $E_{\xi }\uparrow$ & .948 & .958 & .958 & $-$  & .958 & $-$ & .957 & .960 & $-$  & .964 & .971 & \textcolor{red}{\textbf{.974}} & .971 & $-$ & .966 & .969 & \textcolor{blue}{\textbf{.973}}\\
& $M\downarrow$ & .035 & .025 & .035 & .029 & .025 & .027 & .027 & .025 & \textcolor{blue}{\textbf{.020}} & .028 & \textcolor{blue}{\textbf{.020}} & \textcolor{blue}{\textbf{.020}} & \textcolor{red}{\textbf{.019}} & $-$ & .022 & .023 & \textcolor{red}{\textbf{.019}}\\
\hline
\multicolumn{2}{r|}{Params (M)$\downarrow$}  & 63.0 & 47.5 & 58.9 & 103.2  & 31.6 & $-$  & 453.0 & 127.5 & 111.9 & 93.8 & 262.6 & 129.5 & 431.6 & $-$  & 85.4 & 80.0 & 84.5\\
\multicolumn{2}{r|}{FLOPs (G)$\downarrow$} & 126.3 & 44.1 & $-$  & 42.6 & 24.9 & $-$  & 139.7 & 24.1  & 27.0 & 63.9 & 341.8 & 212.0 & 242.2 & $-$  & 40.0 & 49.0 & 18.1\\
\multicolumn{2}{r|}{FPS$\uparrow$}  & 22.0 & 33.4 & $-$  & 54.0  & $-$ & $-$  & 45.0 & $-$  & 21.3 & 26.0 & 11.0 & 23.8 & 9.0 & $-$  & 12.0 & $-$  & 70.2 \\
\hline
	\end{tabular}
\end{table*}

\vspace{1mm}
\noindent \textbf{Selective Interaction Module.}
The structure of SIM is illustrated in Figure~\ref{fig:AFM}~(c). As demonstrated in \cite{dao2024transformers}, Transformers are SSMs, with $\bm{Q}$, $\bm{K}$ and $\bm{V}$ corresponding to $\bm{C}$, $\bm{B}$ and the input $x$. Inspired by the cross-attention mechanism \cite{chen2021crossvit} that interchanges $\bm{Q}$ for information communication, we swap the $\bm{C}$ matrices of RGB and depth features in the process of selective scan for cross-modality interaction. Moreover, the distance maps $D^r$ and $D^d$ are derived by reversing $S^r$ and $S^d$, namely $1-S^r$ and $1-S^d$. By such means, $D^r$ and $D^d$ can focus on the complementary information from the counterparts, and thus are utilized to weight $\bm{C^d}$ and $\bm{C^r}$ respectively. The whole process of SIM can be formulated as follows.
\begin{equation}
\begin{aligned}
    \overline{\bm{A^r}} = \exp(\Delta^r \bm{A^r})&, \ \overline{\bm{A^d}} = \exp(\Delta^d \bm{A^d}),\\
    \overline{\bm{B^r}} = \Delta^r \bm{B^r}&, \ \overline{\bm{B^d}} = \Delta^d \bm{B^d},\\
    h^r_t = \overline{\bm{A^r}} h^r_{t-1} + \overline{\bm{B^r}} X^r_t&, \ h^d_t = \overline{\bm{A^d}} h^d_{t-1} + \overline{\bm{B^d}} X^d_t, \\
    Y^r_t = \textcolor{red}{(D^r\bm{C^d})} h^r_t + \bm{D^r} X^r_t&, \ Y^d_t = \textcolor{red}{(D^d\bm{C^r})} h^d_t + \bm{D^d} X^d_t,
\end{aligned}
\end{equation}
where $\bm{B}$, $\bm{C}$ and $\Delta$ are projected from the input $X$. $\bm{A}$ and $\bm{D}$ are randomly initialized. $t$ denotes the time step.  

\vspace{1mm}
\noindent \textbf{Selective Enhancement Module.}
As shown in Figure~\ref{fig:AFM}~(d), the design philosophy of the SEM is very similar to that of the SIM. To be specific, the similarity maps $S'^r$ and $S'^d$ are separately utilized to weight $\bm{C^{\prime r}}$ and $\bm{C^{\prime d}}$, which enhances the reliability of the single-modality features for subsequent fusion. The process of SEM can be represented as:
\begin{equation}
\begin{aligned}
    \overline{\bm{A^{\prime r}}} = \exp(\Delta^{\prime r} \bm{A^{\prime r}})&, \ \overline{\bm{A^{\prime d}}} = \exp(\Delta^{\prime d} \bm{A^{\prime d}}),\\
    \overline{\bm{B^{\prime r}}} = \Delta^{\prime r} \bm{B^{\prime r}}&, \ \overline{\bm{B^{\prime d}}} = \Delta^{\prime d} \bm{B^{\prime d}},\\
    h^{\prime r}_t = \overline{\bm{A^{\prime r}}} h^{\prime r}_{t-1} + \overline{\bm{B^{\prime r}}} X^{\prime r}_t&, \ h^{\prime d}_t = \overline{\bm{A^{\prime d}}} h^{\prime d}_{t-1} + \overline{\bm{B^{\prime d}}} X^{\prime d}_t, \\
    Y^{\prime r}_t = \textcolor{red}{(S^{\prime r}\bm{C^{\prime r}})} h^{\prime r}_t + \bm{D^{\prime r}} X^{\prime r}_t&, \ Y^{\prime d}_t = \textcolor{red}{(S^{\prime d}\bm{C^{\prime d}})} h^{\prime d}_t + \bm{D^{\prime d}} X^{\prime d}_t.
\end{aligned}
\end{equation}

Thanks to AFM, the overall model can dynamically build up the global relationship between cross-modality features, in turn achieving more comprehensive cross-modality interaction and fusion. 

\section{Experiments}
\subsection{Experimental Setup}

In the following, we state the datasets, evaluation metrics and training implementation. More details can be found in the Appendix.

\vspace{1mm}
\noindent \textbf{Datasets.}
To evaluate the performance of our method, we conduct experiments on seven widely used RGB-D SOD datasets, including NJUD \cite{ju2014depth}, NLPR \cite{peng2014rgbd}, STERE \cite{niu2012leveraging}, SIP \cite{fan2020rethinking}, SSD \cite{zhu2017three}, LFSD \cite{li2014saliency} and DUT-D \cite{piao2019depth}.  
For a fair comparison, we follow the same training settings as \cite{piao2020a2dele, liu2021tritransnet}, where 1485 samples from the NJUD, 700 samples from NLPR and 800 samples from DUT-D are selected as the training set. The rest of NJUD, NLPR and DUT-D, and all of STERE, SIP, SSD, LFSD are used for testing. 

\vspace{1mm}
\noindent \textbf{Evaluation Metrics.}
We adopt four commonly used metrics including F-measure ($F_{\beta}$)~\cite{achanta2009frequency}, S-measure ($S_{\alpha}$)~\cite{fan2017structure}, E-measure ($E_{\xi}$)~\cite{fan2018enhanced} and mean absolute error ($M$) to quantitatively evaluate the performance. For $M$, lower value is better. For others, higher is better. 

\begin{figure*}[t]
  \centering
  \includegraphics[width=0.95\linewidth]{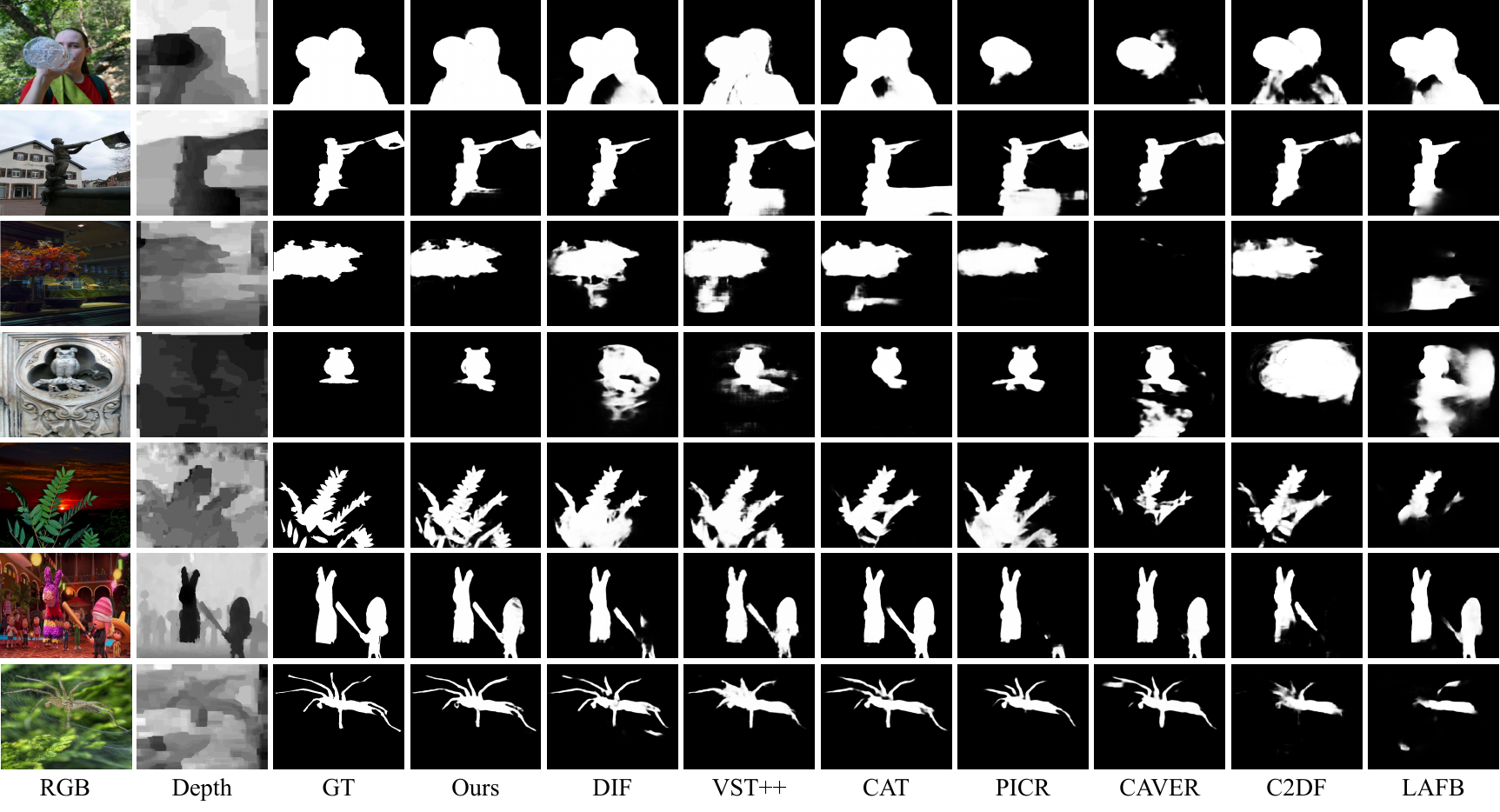}
\vspace{-1mm}
  \caption{Qualitative results of our LEAF-Mamba and other representative methods.}
  \Description{The structure of AFM.}
  \label{fig:compare_sota}
\end{figure*}

\vspace{1mm}
\noindent \textbf{Implementation Details.}
We adopt the VMamba-T \cite{liu2024vmamba} as encoder network, which is pre-trained on ImageNet-1K \cite{krizhevsky2012imagenet}. The network input resolution is $256\times 256$, following \cite{yao2023depth, pang2023caver}. The size of $\{H, W, C\}$ in CSoP is fixed at $\{8, 8, 96\}$. We use pixel position-aware loss \cite{wei2020f3net} for multi-level supervision to pay different attention to the hard and easy pixels. 
Adam \cite{kingma2014adam} algorithm serves as our optimizer.

\subsection{Comparison with State-of-the-arts}
We compare our LEAF-Mamba with recent 16 state-of-the-art methods, in terms of quantitative and qualitative aspects.

\vspace{1mm}
\noindent \textbf{Quantitative evaluation.}
Table~\ref{table:compare with sota} shows the quantitative results of our methods against other 16 state-of-the-art methods on seven benchmark datasets. It can be seen that our network outperforms other advanced models across all the datasets in terms of most evaluation metrics, which demonstrates the superiority of our method in efficacy. For instance, compared with the second best method VST++ \cite{liu2024vst++}, our method improves the $F_{\beta}$ and MAE by 2.38\% and 13.2\% respectively on SSD dataset. 
Meanwhile, we compare our method with other state-of-the-art models in terms of parameters, FLOPs and FPS to evaluate the model size, computation overhead and inference speed, respectively. As shown in the bottom of Table~\ref{table:compare with sota}, the LEAF-Mamba achieves the lowest FLOPs of 18.1G and the highest FPS of 70.2, with comparable parameters of 84.5M, validating its advances in computational efficiency.

\vspace{1mm}
\noindent \textbf{Qualitative evaluation.}
In Figure 5, we visualize some challenging scenes and results generated by our method and other top-ranking models. As we can see, our model can accurately segment objects in varying scales including large object (Row 1), middle object (Row 2-3), small object (Row 4) and multiple objects (Row 6), which demonstrates the effectiveness of our LE-SSM in extracting multi-scale information. Meanwhile, for objects with complex backgrounds (Row 3 and 7) or low-quality depth maps (Row 4), our method is able to generate fine predictions that are more consistent with the ground truth, benefiting from the promising cross-modality interaction achieved by our proposed AFM.  

\newcommand{\hlsim}[1]{%
  \colorbox{colorsim}{\rule[-0.1ex]{0pt}{0.5ex}#1}%
}

\newcommand{\baseline}[1]{\colorbox{colorbaseline}{#1}}
\newcommand{\hlsem}[1]{\colorbox{colorist}{#1}}
\newcommand{\hlours}[1]{\colorbox{colorleaf}{#1}}

\begin{table}[t]

  \caption{Ablation analyses of each component on the NJUD and SSD datasets. bold: top-1 results.}
    \vspace{-1mm}
  \label{tab:ablation_all}
  \small
  \tablestyle{6pt}{\lineheight}
  \begin{tabular}{ccccccccc}
			\toprule
			\multirow{2}{*}{No.} &\multirow{2}{*}{Configuration}   & \multicolumn{2}{c}{\textit{NJUD}}	&\multicolumn{2}{c}{\textit{SSD}} 	  \\
			\cmidrule(r){3-4}       \cmidrule(r){5-6}
			{~}                          &{~}               &$F_{\beta}$ $\uparrow$    &$M$ $\downarrow$              &$F_{\beta}$ $\uparrow$    &$M$ $\downarrow$      \\       
			\midrule
			\#1              & Baseline     & \baseline{.917}    & \baseline{.030}             & \baseline{.851}    & \baseline{.044}                       \\		
			\#2               & + LE-SSM                & \hlsim{.931}   & \hlsim{.028}             & \hlsim{.872}    & \hlsim{.039}        \\
			\#3             & + AFM      &\hlsem{.940}    & \hlsem{.026}             &\hlsem{.891}    & \hlsem{.035}                    \\
			\#4        & + LE-SSM + AFM (LEAF)    &\hlours{\textbf{.945}}    & \hlours{\textbf{.025}}          &\hlours{\textbf{.904}}    &\hlours{\textbf{.033}}            \\
			
			\bottomrule
\end{tabular}
\end{table}

\subsection{Ablation Studies}
We conduct ablation studies to verify the effectiveness of two main components (LE-SSM and AFM) in the LEAF-Mamba on the NJUD and SSD datasets, and choose $F_{\beta}$ and MAE for evaluation. The quantitative results are summarized in Table~\ref{tab:ablation_all}.

\vspace{1mm}
\noindent \textbf{Components ablation.}
In this part, we firstly evaluate the effectiveness of the key components (namely, LE-SSM and AFM) in our LEAF-Mamba. Results are reported in Table~\ref{tab:ablation_all}. Our baseline (No. \#1) employs a two-stream VMamba-based network where the dual-modality features with the same resolution are added directly. Detailed baseline architecture is presented in the Appendix. Compared with baseline, our LE-SSM/AFM obtains numerical improvements, e.g., 2.47\%$/$4.7\% and 11.4\%$/$20.0\% increase on SSD in terms of $F_{\beta}$ and MAE, respectively. Furthermore, when concurrently adopting LE-SSM and AFM, our LEAF-Mamba raises the gains to 6.23\% and 25\%, showing the synergy between the two modules.

\vspace{1mm}
\noindent \textbf{Scanning strategy in LE-SSM.}
We assess the effectiveness of various scanning strategies in LE-SSM. Specifically, we compare our MSW-SS2D with SS2D~\cite{liu2024vmamba}, continuous scan~\cite{yang2024plainmamba} and fixed windowed scan~\cite{huang2024localmamba}. The detailed illustration of them can be found in the Appendix. The results are presented in Table~\ref{table:ablition_scan}.  It can be observed that, our MSW-SS2D strategy surpasses all counterparts with a clear margin, demonstrating the positive gains of comprehensively modeling local dependencies in a multi-scale manner.

\begin{table}[t]	
    
	\caption{Quantitative results of various scanning strategies for local enhancement (LE). bold: top-1 results.}
    \vspace{-1mm}
	\centering
    \small
    \tablestyle{4pt}{\lineheight}
	\label{table:ablition_scan}
		\begin{tabular}{cccccc}
			\toprule
			\multirow{2}{*}{No.}	&\multirow{2}{*}{Scanning  Strategy}      & \multicolumn{2}{c}{\textit{NJUD}}	&\multicolumn{2}{c}{\textit{SSD}} 	  \\
			\cmidrule(r){3-4}       \cmidrule(r){5-6}
			{~}                            &{~}        &$F_{\beta}$ $\uparrow$    &$M$ $\downarrow$              &$F_{\beta}$ $\uparrow$    &$M$ $\downarrow$    \\       
			\midrule
            \#1 & SS2D & \baseline{.917}    & \baseline{.030}             & \baseline{.851}    & \baseline{.044}                       \\
			\#5		 &  Continuous scan \cite{yang2024plainmamba}                                            &.919    &.030             &.856    &.043                    \\
			\#6		 & Fixed windowed scan \cite{huang2024localmamba}                                       &.925    &.029             &.863    &.041                   \\
			\#2		 &MSW-SS2D (Ours)                               & \hlsim{\textbf{.931}}    &\hlsim{\textbf{.028}}          &\hlsim{\textbf{.872}}    & \hlsim{\textbf{.039}}            \\
			
			\bottomrule
	\end{tabular}
\end{table}

\begin{table}[t]
\tablestyle{6pt}{1.2}
  \caption{Ablation analyses of AFM. bold: top-1 results.}
  \label{tab:ablation_AFM}
  \begin{tabular}{lcccccccc}
			\toprule
			\multirow{2}{*}{No.}  & \multirow{2}{*}{CSoP} & \multirow{2}{*}{SIM} & \multirow{2}{*}{SEM} & \multicolumn{2}{c}{\textit{NJUD}}	&\multicolumn{2}{c}{\textit{SSD}} 	  \\
            \cmidrule(r){5-6}       \cmidrule(r){7-8}
             & & & &$F_{\beta}$ $\uparrow$    &$M$ $\downarrow$              &$F_{\beta}$ $\uparrow$    &$M$ $\downarrow$      \\       
            \midrule
            \#1                   &          &          &                  & \baseline{.917}    & \baseline{.030}             & \baseline{.851}    & \baseline{.044}                       \\
            \midrule
            \#7 &  &  $\checkmark$ & &.924 &.029 &.863 &.042 \\
            \#8  &  &  &$\checkmark$ &.921 &.030 &.858 &.043\\
            \#9 & & $\checkmark$ &  $\checkmark$ &.927 &.028 &.867 &.041 \\
            \midrule
            \#10 & $\checkmark$ &$\checkmark$ & &.933 &.027 &.879 &.038 \\
            \#11 &$\checkmark$  &  &$\checkmark$ &.926 &.028 &.872 &.040\\
            \#3  &$\checkmark$  &  $\checkmark$ &$\checkmark$ & \hlsem{\textbf{.940}} & \hlsem{\textbf{.026}} & \hlsem{\textbf{.891}} & \hlsem{\textbf{.035}} \\
			\bottomrule
\end{tabular}
\end{table}

\vspace{1mm}
\noindent \textbf{Design of AFM.}
In the following, we evaluate the effect of core components in our AFM, i.e., CSoP, SIM and SEM. The results are reported in Table~\ref{tab:ablation_AFM}. In the absence of CSoP, independently applying the SIM (or SEM) achieves performance boost with 1.41\% (0.82\%) and 4.54\% (2.27\%) on the SSD dataset in terms of $F_{\beta}$ and $M$ metrics. Moreover, adopting them together yields impressive performance gains of 1.88\% and 6.81\%. These observations admit the effectiveness of SSM mechanism in dual-modality interaction and single-modality enhancement. Based on these findings, we conduct in-depth exploration on the role of CSoP-based selective mechanism for SIM and SEM, respectively (No. \#10 \& \#11). All of these results typically make further improvements, benefiting from the adaptive weights from CSoP. Ultimately, our AFM (with CSoP, SIM and SEM) achieves the best performance. More detailed empirical studies on CSoP, SIM and SEM are provided in the Appendix.

\subsection{Application to RGB-T SOD}
To validate the generalization ability of LEAF-Mamba, we extend it to the RGB-Thermal (RGB-T) SOD task and conduct experiments on three public RGB-T SOD datasets, i.e., VT821 \cite{wang2018rgb}, VT1000 \cite{tu2019rgb}, and VT5000 \cite{tu2022rgbt}. Following \cite{tu2021multi, pang2023caver}, the training set contains 2500 images from VT5000, with the remaining images used for testing. 

\begin{figure}[t]
  \centering
  \includegraphics[width=0.98\linewidth]{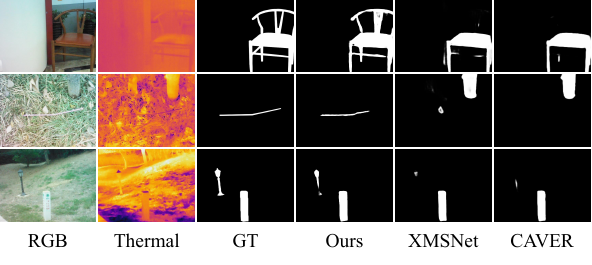}
  \vspace{-2mm}
  \caption{Visual comparisons on RGB-T SOD.}
  \Description{The structure of AFM.}
  \label{fig:visual_comparison_rgbt}
  \vspace{-2mm}
\end{figure}

We compare our network with 6 recent RGB-T SOD methods and show the quantitative results in Table~\ref{table:quantitative_comparison_rgbt}. It can be seen that our model achieves overall competitive performance on the three benchmarks, e.g., with an improvement of 1.20\% and 9.09\% on VT821 in term of $S_{\alpha}$ and MAE. Additionally, the visual comparisons shown in Figure~\ref{fig:visual_comparison_rgbt} indicate that LEAF-Mamba excels in challenging scenarios such as low-quality thermal images, complex backgrounds and multi-scale objects, further demonstrating its effectiveness and prominent generalizability on multi-modality tasks.

\begin{table}[t]
    
    \tablestyle{3.5pt}{1.1}

	\caption{Quantitative comparisons with recent RGB-T SOD methods on three benchmarks. The best two results are shown in \textcolor{red}{Red} and \textcolor{blue}{Blue}. $-$ means not avaliable.}
	\centering
	\label{table:quantitative_comparison_rgbt}
		\begin{tabular}{lrccccccc}
           \toprule

\multicolumn{2}{c}{\multirow{2}{*}{Method}}  &  \ourfont{CAVER} & \ourfont{LSNet} & \ourfont{CMDBIF} & \ourfont{XMSNet} & \ourfont{LAFB} & \ourfont{ConTriNet} & \ourfont{LEAF} \\

& & \cite{ji2022dmra} & ~\cite{zhou2023lsnet} & \cite{xie2023cross} & \cite{wu2023object} & \cite{wang2024learning} & ~\cite{tang2024divide} & (Ours) \\

\multicolumn{2}{c}{Publish} & \ourfont{TIP} & \ourfont{TIP} & \ourfont{TCSVT} & \ourfont{MM} & \ourfont{TCSVT} & \ourfont{TPAMI} & -- \\
\multicolumn{2}{c}{Year}   & 2023 & 2023 & 2023 & 2023 & 2024 & 2024 & -- \\


\midrule
\multirow{4}{*}{\rotatebox{90}{\textit{VT821}}} &$F_{\beta}\uparrow$  & .877     &.827    &.855    &.859      &.843     &\textcolor{blue}{\textbf{.878}}      &\textcolor{red}{\textbf{.885}}\\
 & $S_{\alpha }\uparrow$ & .898    &.877     &.882    &.906        &$-$    &\textcolor{blue}{\textbf{.915}}   &\textcolor{red}{\textbf{.926}}\\
& $E_{\xi }\uparrow$ & .928   &.911     &.927    &.929         &.915   &\textcolor{blue}{\textbf{.940}}  &\textcolor{red}{\textbf{.943}}\\
&$M\downarrow$& .027   &.033     &.032    &.028         &.043  &\textcolor{blue}{\textbf{.022}} &\textcolor{red}{\textbf{.020}}\\
\midrule

\multirow{4}{*}{\rotatebox{90}{\textit{VT1000}}} & $F_{\beta}\uparrow$ & \textcolor{red}{\textbf{.939}}    &.887    &.914    &.903          &.905   &.918  &\textcolor{blue}{\textbf{.926}}\\
& $S_{\alpha }\uparrow$ & .938    &.924     &.927    &.936             &$-$  &\textcolor{blue}{\textbf{.941}}  &\textcolor{red}{\textbf{.945}}\\
& $E_{\xi }\uparrow$ & .949   &.936     &\textcolor{red}{\textbf{.967}}    &.945             &.945    &.954 
 &\textcolor{blue}{\textbf{.962}}\\
&$M\downarrow$& \textcolor{blue}{\textbf{.017}}    &.022     &.019    &.018        &.018  &\textcolor{red}{\textbf{.015}}  &\textcolor{red}{\textbf{.015}}\\

\midrule

\multirow{4}{*}{\rotatebox{90}{\textit{VT5000}}} & $F_{\beta}\uparrow$ & .882    &.827    &.869    &.871       &.857      &\textcolor{red}{\textbf{.898}}  &\textcolor{blue}{\textbf{.893}}\\  
&  $S_{\alpha }\uparrow$ &.899    &.876     &.886    &.907      &$-$       &\textcolor{red}{\textbf{.923}} &\textcolor{blue}{\textbf{.919}}\\
& $E_{\xi }\uparrow$ & .941    &.916     &.937    & .939     &.931   &\textcolor{blue}{\textbf{.956}}  &\textcolor{red}{\textbf{.958}}\\
&$M\downarrow$ & .028    &.036     &.032    &.028      &.030      &\textcolor{red}{\textbf{.020}} &\textcolor{blue}{\textbf{.021}}\\

\bottomrule
\end{tabular}
\end{table}

\section{Conclusion}
In this paper, we propose a novel SSM-based system (LEAF-Mamba) to address both efficacy and efficiency bottlenecks in existing RGB-D salient object detection. We develop a local emphatic state space module equipped with the multi-scale windowed scanning strategy to capture multi-scale local dependencies in the process of feature extraction. We also design an SSM-based adaptive fusion module to dynamically select the discriminative regions of two modalities for complementary cross-modality interaction, as well as the similar ones for reliable cross-modality fusion. Experimental results on seven benchmarks show superior performance of our method over 16 state-of-the-art models in both accuracy and efficiency. Furthermore, extensive ablation studies demonstrate the effectiveness of each proposed component.

\section{Acknowledgments}

This work is supported by the Ministry of Education, Singapore, under its MOE AcRF Tier 3 Grant (MOE-MOET32022-0001).

\balance
\putbib[refs]
\end{bibunit}

\clearpage
\appendix

\begingroup
\makeatletter

\renewcommand{\bibnumfmt}[1]{[R#1]}
\def\NAT@open{[R}
\def\NAT@close{]}

\@ifpackageloaded{hyperref}{%
  \let\orig@lbibitem\@lbibitem
  \def\@lbibitem[#1]#2{%
    \Hy@raisedlink{\hyper@anchorstart{cite.app-#2}\hyper@anchorend}%
    \orig@lbibitem[{#1}]{#2}%
  }%
  \let\orig@bibitem\@bibitem
  \def\@bibitem#1{%
    \Hy@raisedlink{\hyper@anchorstart{cite.app-#1}\hyper@anchorend}%
    \orig@bibitem{#1}%
  }%
  \let\orig@hyper@natlinkstart\hyper@natlinkstart
  \def\hyper@natlinkstart#1{\orig@hyper@natlinkstart{app-#1}}%
}{}

\setcounter{figure}{0}\setcounter{table}{0}
\renewcommand{\thefigure}{F\arabic{figure}}
\renewcommand{\thetable}{T\arabic{table}}

\makeatother

\begin{bibunit}

\noindent \textbf{\Large{Appendix}}
\vspace{1mm}

\noindent This Appendix provides additional details and results that complement the main manuscript, which are omitted due to page limitations. The contents are organized as follows:
\begin{itemize}
\item Dataset Specification in \S~\ref{sec:datasets};  
\item Metrics Description in \S~\ref{sec:metric};  
\item Experimental Implementation in \S~\ref{sec:impl.};  
\item Architecture of Baseline in \S~\ref{sec:extra_bs};
\item Scanning Strategy in \S~\ref{sec:scan_path};
\item Additional Experiments in \S~\ref{sec:extra_exp};
\item Feature Visualization in \S~\ref{sec:visualization}.
\item Failure Cases in \S~\ref{sec:failure_case}.
\end{itemize}

\section{Datasets}\label{sec:datasets}

We conduct extensively experiments on ten multi-modality salient object detection (SOD) datasets, including seven RGB-Depth (RGB-D) benchmarks and three RGB-Thermal (RGB-T) ones. The prior datastes includes NJUD~\cite{ju2014depth}, NLPR~\cite{peng2014rgbd}, STERE~\cite{niu2012leveraging}, SIP~\cite{fan2020rethinking}, SSD~\cite{zhu2017three}, LFSD~\cite{li2014saliency}, and DUT-D~\cite{piao2019depth}; and the latter ones involve VT821 \cite{wang2018rgb}, VT1000 \cite{tu2019rgb} and VT5000 \cite{tu2022rgbt}. The characteristics of each dataset are summarized below.

\noindent \textbf{RGB-D Datasets.}
NJUD~\cite{ju2014depth} contains 1,985 pairs of RGB and depth images collected from the Internet, 3D movies, and stereo photographs. Dataset NLPR~\cite{peng2014rgbd} includes 1,000 RGB-D pairs covering a variety of indoor and outdoor scenes. STERE~\cite{niu2012leveraging} consists of 1,000 stereoscopic images sourced from Flickr, NVIDIA 3D Vision Live, and the Stereoscopic Image Gallery. SIP~\cite{fan2020rethinking} is a high-resolution dataset comprising 929 image pairs captured in outdoor environments with complex lighting and diverse human poses. SSD~\cite{zhu2017three} includes 80 samples from both indoor and outdoor scenes. LFSD~\cite{li2014saliency} provides 100 RGB-D image pairs designed for saliency detection in light field images. DUT-D~\cite{piao2019depth} consists of 1,200 RGB-D pairs, with 800 captured indoors and 400 captured outdoors.
We follow the training protocol as in previous works~\cite{piao2019depth, piao2020a2dele} and \cite{liu2021tritransnet}. Specifically, 1,485 samples from NJUD, 700 samples from NLPR, and 800 samples from DUT-D are used for training. The remaining samples from NJUD, NLPR, and DUT-D, along with all samples from STERE, SIP, SSD, and LFSD, are used for testing.

\noindent \textbf{RGB-T Datasets.} 
VT821~\cite{wang2018rgb} contains 821 manually aligned RGB-T image pairs; VT1000~\cite{tu2019rgb}, comprising 1,000 image pairs captured in relatively simple scenes with well-aligned sensors; and VT5000~\cite{tu2022rgbt}, which includes 5,000 high-resolution and diverse image pairs with minimal misalignment.
Following the training protocol of \cite{tu2021multi, pang2023caver}, we use 2,500 image pairs from VT5000 for training, while the remaining samples from VT5000, along with all images from VT821 and VT1000, are used for evaluation.

\section{Evaluation Metrics}\label{sec:metric}

\newcommand{\todo}[1]{\textcolor{red}{#1}}

We employ four popular metrics to assess the performance, following \cite{sun2023catnet, chen2024trans}. F-measure ($F_{\beta}$)~\cite{achanta2009frequency} is a region-based similarity metric based on precision and recall. S-measure ($S_{\alpha}$)~\cite{fan2017structure} focuses on region-aware and object-aware structural similarities between the saliency map and the ground truth. E-measure ($E_{\xi}$)~\cite{fan2018enhanced} is characterized as both image-level statistics and local pixel matching. Mean absolute error (MAE, $M$) measures the average difference between the prediction and the ground truth in the pixel level. The lower value is better for $M$ and the higher is better for others.

\section{Training Implementation}\label{sec:impl.}

Our proposed LEAF-Mamba is implemented using the PyTorch toolkit and trained on a PC equipped with a single NVIDIA RTX 4090 GPU. VMamba-T~\cite{liu2024vmamba} is employed as the encoder network, initialized with weights pre-trained on ImageNet-1K~\cite{krizhevsky2012imagenet}. The size of $\{H, W, C\}$ in the CSoP module is fixed at $\{8, 8, 96\}$. Following~\cite{yao2023depth, pang2023caver}, all images are uniformly resized to 256$\times$256 during both training and inference. During training phase, random flipping and random rotation are employed for data augmentation to alleviate overfitting. The Adam algorithm \cite{kingma2014adam} serves as the optimizer with a mini-batch size of 8. The initial learning rate is set to 1e-4 and decayed by a factor of 10 every 60 epochs. The training process runs for a total of 200 epochs.

\begin{figure*}[t]
  \centering
  \includegraphics[width=0.8\linewidth]{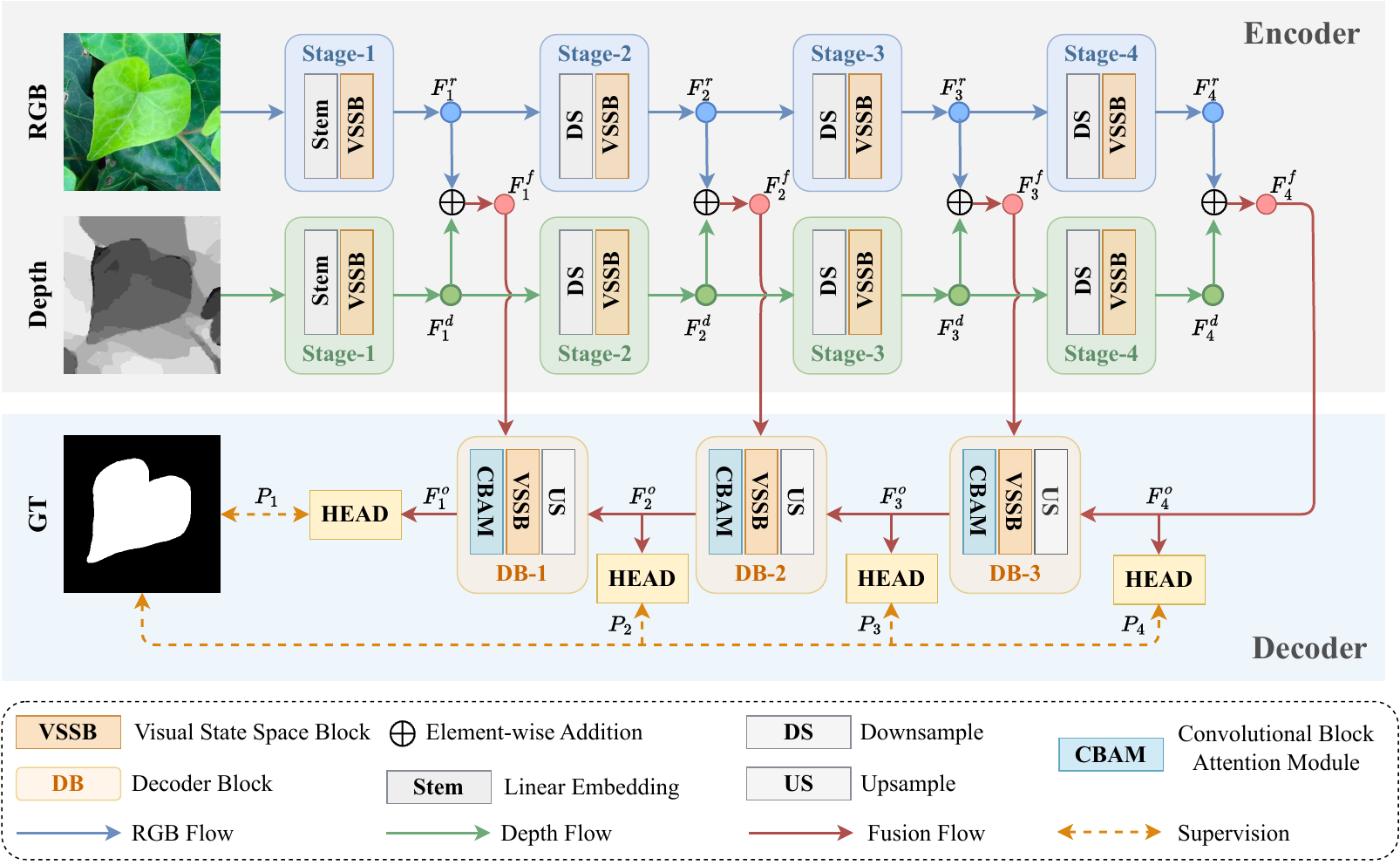}
  \caption{Architecture of the baseline.}
  \Description{Baseline.}
  \label{fig:baseline}
\end{figure*}

\begin{figure*}[t]
  \centering
  \includegraphics[width=0.72\linewidth]{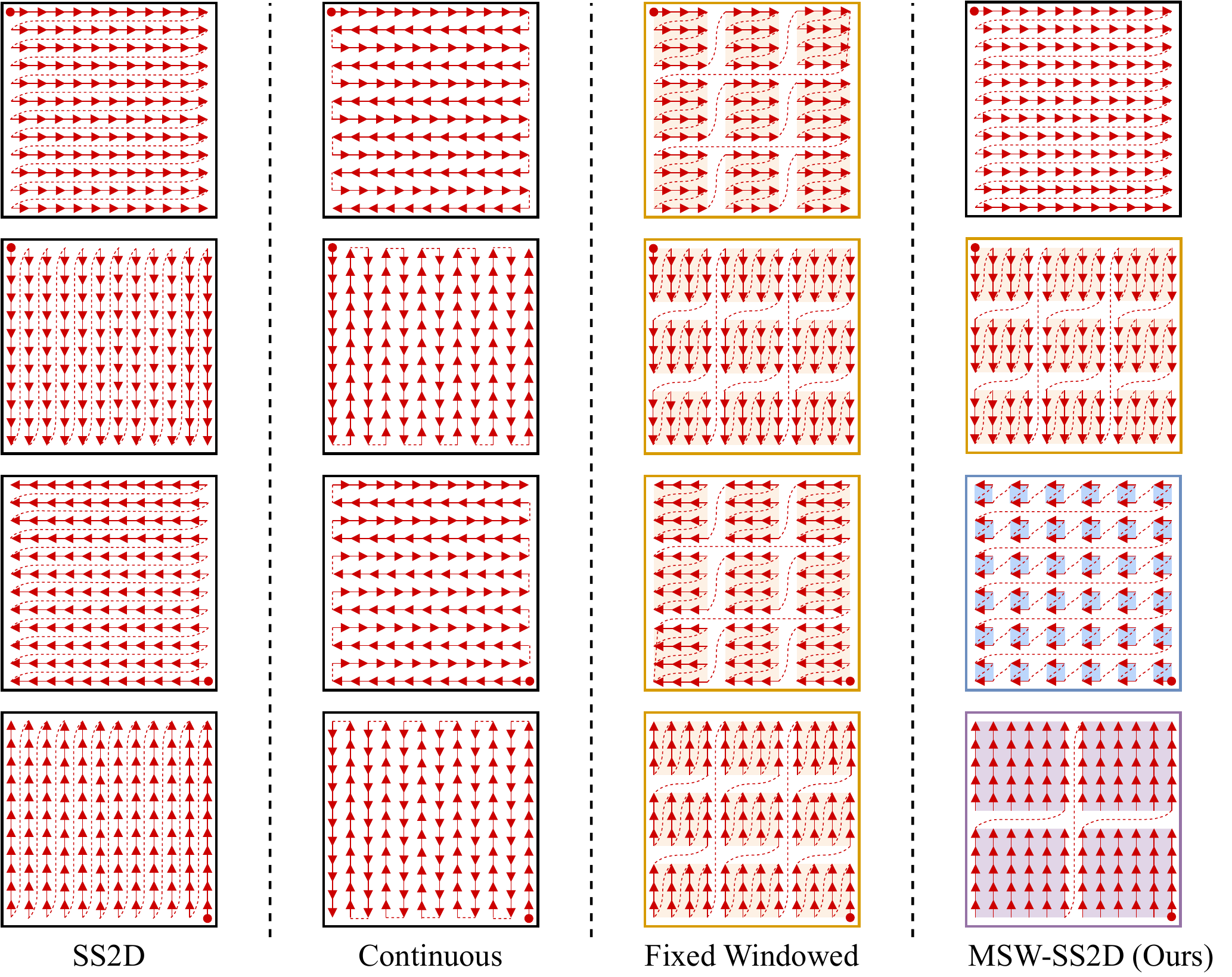}
  \caption{Various scanning strategy counterparts.}
  \Description{scan.}
  \label{fig:scan_path}
\end{figure*}

\newcommand{\hlours}[1]{\colorbox{colorleaf}{#1}}

\newcommand{\hlsim}[1]{%
  \colorbox{colorsim}{\rule[-0.1ex]{0pt}{0.5ex}#1}%
}

\newcommand{\baseline}[1]{\colorbox{colorbaseline}{#1}}
\newcommand{\hlsem}[1]{\colorbox{colorist}{#1}}

\newcommand{\tablestyle}[2]{\setlength{\tabcolsep}{#1}\renewcommand{\arraystretch}{#2}\centering\small}

\newcommand{\oldid}[1]{\textcolor{gray}{#1}}

\section{Architecture of Baseline}\label{sec:extra_bs}
The detailed architecture of baseline is illustrated in Figure~\ref{fig:baseline}. We adopt the two-stream VMamba-T as encoder which is initialized by the parameters pre-trained on ImageNet-1K. The RGB and depth features with the same resolution are added directly for cross-modality fusion to generate RGB-D features. The multi-stage RGB-D features are delivered into an SSM-based FPN-like decoder for prediction. Each decoder block contains one upsample layer, one VMamba block and one CBAM.

\section{Scanning Strategy}\label{sec:scan_path}

We illustrate various scanning strategy counterparts in Figure~\ref{fig:scan_path}. As shown in figure, SS2D~\cite{liu2024vmamba} performs the row-wise and column-wise scanning along the same direction. Different from SS2D, continuous scan~\cite{yang2024plainmamba} reverses the scanning direction in adjacent lines, leading to a Zigzag scanning path, which partially reserves the local proximity. In addition, fixed windowed scan~\cite{huang2024localmamba} splits the feature map into several local regions and performs SS2D-like scanning in each window. In contrast to these previous works, our MSW-SS2D concurrently considers multi-scale semantics in once four-way scanning. Therefore, our MSW-SS2D is potential to capture multi-scale local dependencies. Equipped with the MSW-SS2D, our LEAF-Mamba achieves superior performance on both the quantitative and qualitative experiments in Table 3 and Figure 5 of the manuscript, demonstrating its advance in detecting multi-scale salient objects.

\section{Additional Experiments}\label{sec:extra_exp}

In this section, we elaborately validate the effects of some components in our method, including CSoP, SIM and SEM. The experiments are conducted on NJUD and SSD datasets and $F_{\beta}$ and $M$ are chosen for evaluation. The identical configuration are denoted in the same color. The best result is highlighted in \textbf{bold}.

\subsection{Ablation Study on CSoP}
\vspace{1mm}
\noindent \textbf{Similarity.}
Initially, we evaluate the devise of CSoP, including the similarity function and pooling approach. Results are presented in Table~\ref{tab:ablation_csop}. In the first pannel, we compare two similarity implementation: a point-to-point cosine similarity (No. \#A2) and a global cross-covariance matrix (No. \#A3, indicating Eqn. (7) in manuscript). To be specific, for the RGB token and depth token $T^r\in\mathbb{R}^{HW\times C}$ and $T^d\in\mathbb{R}^{HW\times C}$, the cosine similarity between $T^r_i$ and $T^d_i$ can be written as:
\begin{equation}
  M_{i}^{Cos} = \mathcal{F}\left(T_i^{d}, T_i^r\right) = \frac{\left<T_i^{d}, T_i^r\right>} {\|T_i^{d}\|\cdot \|T_i^r\|},.
\end{equation}
where $\mathcal{F}\left(\cdot\right)$ denotes the cosine similarity function, notation $\left<\cdot,\cdot\right>$ and $\|\cdot\|$ indicate inner production and $\ell_2$ normalization operation respectively.
And the cross-covariance similarity for $T^r_i$ and $T^d_j$ can be formulated as: 
\begin{equation}
   M_{i,j}^{Cov} = \mathcal{H}\left(T^{d}_{i}, T^{r}_{j}\right) = \frac{1}{C-1}\left<T_i^{d} - \bar{T}^{d}_{i}, T_j^r - \bar{T}^{r}_{j}\right>
\end{equation}
\begin{equation}
   \bar{T}^{d}_{i} = \frac{1}{C} \sum_{c=1}^{C} T_{i,c}^{d},~~\bar{T}^{r}_{j} = \frac{1}{C} \sum_{c=1}^{C} T_{j,c}^{r},. 
\end{equation}
where notation $\mathcal{H}\left(\cdot\right)$ denotes the cross-covariance similarity function.
From the results, we can observe that, considering similarity in CSoP obtains performance gains over baseline (No. \#A1). In addition, covariance similarity is superior to the cosine one with an advance of 1.59\% and 7.89\% on $F_{\beta}$ and $M$ metric of SSD dataset. Thus, covariance similarity is used as the default setting in the CSoP.

\vspace{1mm}
\noindent \textbf{Pooling.} In the second pannel of Table~\ref{tab:ablation_csop}, we assess various pooling functions for the covariance similarity matrix $M^{Cov}$. Among all pooling implementations, the convolutional pooling achieves the best performance. It leads average$/$max pooling over 1.25\%$/$0.91\% and 7.89\%$/$0.54\% on $F_{\beta}$ and $M$. Due to its superiority, convolution pooling serves as the default configuration for CSoP.

\begin{table}[t]
  \caption{Ablation analyses of CSoP module on the NJUD and SSD datasets. bold: top-1 results. Numbers shown in \oldid{gray} indicates to those referenced in the manuscript.}
  \label{tab:ablation_csop}
  \small
  \tablestyle{6pt}{1.2}
  \begin{tabular}{lcccccc}
			\toprule
    \multirow{2}{*}{No.}  &\multirow{2}{*}{Similarity}   & \multicolumn{2}{c}{\textit{NJUD}}	&\multicolumn{2}{c}{\textit{SSD}} 	  \\
    \cmidrule(r){3-4}       \cmidrule(r){5-6}
    {~}                                &{~}               &$F_{\beta}$ $\uparrow$    &$M$ $\downarrow$              &$F_{\beta}$ $\uparrow$    &$M$ $\downarrow$      \\       
    \midrule
    \#A1 \oldid{(\#1)}  &$-$   & \baseline{.917} & \baseline{.030} & \baseline{.851}  & \baseline{.044} \\
    \#A2 & Cosine &.932 &.027 &.877  &.038\\
    \#A3 \oldid{(\#3)} & Covariance (AFM) &  \hlsem{\textbf{.940}} & \hlsem{\textbf{.026}} & \hlsem{\textbf{.891}} & \hlsem{\textbf{.035}} \\
    \midrule
   \multirow{2}{*}{No.} &\multirow{2}{*}{Pooling}   & \multicolumn{2}{c}{\textit{NJUD}}	&\multicolumn{2}{c}{\textit{SSD}} 	  \\
     \cmidrule(r){3-4}       \cmidrule(r){5-6}
{~}                          &{~}               &$F_{\beta}$ $\uparrow$    &$M$ $\downarrow$              &$F_{\beta}$ $\uparrow$    &$M$ $\downarrow$      \\  

\midrule
\#A1 \oldid{(\#1)} & $-$ & \baseline{.917} & \baseline{.030} & \baseline{.851}  & \baseline{.044} \\
\#A4 & AvgPool &.933 &.027 &.880 &.038\\
\#A5 & MaxPool &.935 &.027 &.883 &.037\\
\#A3 \oldid{(\#3)} &  ConvPool (AFM) &  \hlsem{\textbf{.940}} & \hlsem{\textbf{.026}} & \hlsem{\textbf{.891}} & \hlsem{\textbf{.035}}  \\
			
\bottomrule
\end{tabular}
\end{table}

\begin{table}[t]

  \caption{Ablation analyses of swap feature in SIM on the NJUD and SSD datasets. bold: top-1 results. Numbers shown in \oldid{gray} indicates to those referenced in the manuscript.}
  \label{tab:ablation_sim}
  \small
  \tablestyle{4pt}{1.2}
  \begin{tabular}{lccccccc}
	\toprule
    \multirow{2}{*}{No.} & \multirow{2}{*}{\makecell{Cross-modality\\Interaction}}& \multirow{2}{*}{Swap}   & \multicolumn{2}{c}{\textit{NJUD}}	&\multicolumn{2}{c}{\textit{SSD}} 	  \\
    \cmidrule(r){4-5}       \cmidrule(r){6-7}
    {~}                    &{~}   &            &$F_{\beta}$ $\uparrow$    &$M$ $\downarrow$              &$F_{\beta}$ $\uparrow$    &$M$ $\downarrow$      \\       
    \midrule
    \#A1 \oldid{(\#1)} & $\times$ & $-$ & \baseline{.917} & \baseline{.030} & \baseline{.851}  & \baseline{.044}   \\
    \#A6 & $\checkmark$ & $\mathit{X}$ &.919 &.030  &.855 &.044\\
    \#A7 & $\checkmark$ & $\bm{B}$ &.921 &\textbf{.029}  &.857 &.043\\
    \#A8 \oldid{(\#7)} & $\checkmark$ & $\bm{C}$ & \textbf{.924} & \textbf{.029} & \textbf{.863} & \textbf{.042} \\

 \bottomrule
\end{tabular}
\end{table}

\subsection{Ablation Study on SIM}

In this part, we evaluate the devise of SIM. To be specific, we compare the various swap objects in cross-modality interaction operation, namely $X$, $\bm{B}$ and $\bm{C}$. As shown the results in Table~\ref{tab:ablation_sim}, considering the cross-modality interaction (No. \#A6\&A7\&A8) obtains novel performance gains over baseline (No. \#A1). Moreover, among all cross-modality interaction implementation, swapping $\bm{C}$ obtains the best performance. Based on these findings, our SIM module select $\bm{C}$ to perform modality knowledge promotion.

\subsection{Ablation Study on SEM}

In addition, we consider various weighting objects in SEM. We report the results of weighting various candidates ($X$, $\bm{B}$ and $\bm{C}$) in Table~\ref{tab:ablation_sem}. As we can see, considering the different roles and reliability of each features generally performs better than the equally regarding approach (baseline, No. \#A1). Among various weighting options, $\bm{C}$ lags its counterparts $X$/$\bm{B}$ with a clear margin of 0.93\%/0.58\% and  4.76\%/2.44\% in $F_{\beta}$ and $M$ on the SSD dataset. Accordingly, weighting on $\bm{C}$ is set to the default configuration in our SEM module.

\begin{table}[t]

  \caption{Ablation analyses of weighting object in SEM on the NJUD and SSD datasets. bold: top-1 results.  Numbers shown in \oldid{gray} indicates to those referenced in the manuscript.}
  \label{tab:ablation_sem}
  \small
  \tablestyle{8pt}{1.2}
  \begin{tabular}{lcccccc}
	\toprule
    \multirow{2}{*}{No.} & \multirow{2}{*}{Weighting}   & \multicolumn{2}{c}{\textit{NJUD}}	&\multicolumn{2}{c}{\textit{SSD}} 	  \\
    \cmidrule(r){3-4}       \cmidrule(r){5-6}
    {~}                    &{~}               &$F_{\beta}$ $\uparrow$    &$M$ $\downarrow$              &$F_{\beta}$ $\uparrow$    &$M$ $\downarrow$      \\       
    \midrule
    \#A1 \oldid{(\#1)} & $-$ & \baseline{.917} & \baseline{.030} & \baseline{.851}  & \baseline{.044}   \\
    \#A9 & $\mathit{X}$ & .925 &\textbf{.028} &.864 &.042 \\
    \#A10 & $\bm{B}$ &.923 &.029 &.867 &.041\\
    \#A11 \oldid{(\#11)} & $\bm{C}$ &  \textbf{.926} & \textbf{.028} & \textbf{.872} & \textbf{.040} \\		
	\bottomrule
\end{tabular}
\end{table}

\begin{figure}[t]
  \centering
  \includegraphics[width=0.98\linewidth]{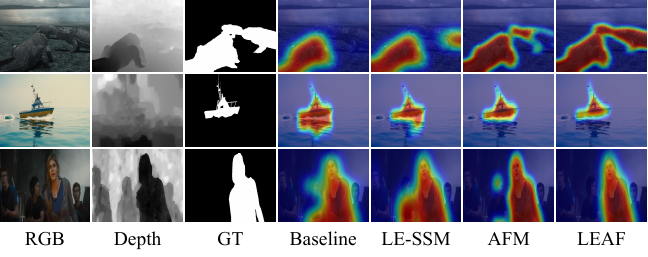}
  \caption{Visualization analyses of the LEAF-Mamba.}
  \Description{The structure of AFM.}
  \label{fig:heatmap}
\end{figure}

\begin{figure}[t]
  \centering
  \includegraphics[width=0.9\linewidth]{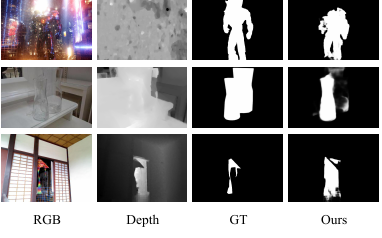}
  \caption{Failure cases illustration.}
  \Description{Task comparisons.}
  \label{fig:failure_case}
\end{figure}

\section{Feature Visualization}\label{sec:visualization}

Intuitively, we visualize some feature maps with different configurations in Figure~\ref{fig:heatmap}. As we can see, compared with the baseline, the model with LE-SSM pays more attention to the multi-scale objects (Row 1) and presents better edge details (Row 3), which verifies the effectiveness of LE-SSM in modeling multi-scale local dependencies. With the incorporation of AFM, RGB and depth modalities exhibit improved collaboration. Concretely, for complex background (Row 3) or low-quality depth (Row 2), AFM can make full use of the complementary cue in both modalities as well as exclude their unreliable content, thus contributing to a more accurate prediction. With LE-SSM and AFM combined, our LEAF-Mamba can focus on the salient objects with the consistency of the ground truth.

\section{Failure Cases}\label{sec:failure_case}

To provide a more comprehensive evaluation of LEAF-Mamba, we present several failure cases in Fig.\ref{fig:failure_case}. In the first row, when both RGB and depth inputs are of low quality, the model struggles to extract insightful knowledge from either modality, resulting in inaccurate predictions. Moreover, certain challenging scenes also present inherent difficulties, such as transparent glass in row \#2 and a partially occluded kite in row~\#3.

\balance
\putbib[refs_app]  

\end{bibunit}
\endgroup

\end{document}